\documentclass[lettersize,journal]{IEEEtran}
\usepackage{amsmath,amsfonts}
\usepackage{algorithmic}
\usepackage{algorithm}
\usepackage{array}
\usepackage[caption=false,font=normalsize,labelfont=sf,textfont=sf]{subfig}
\usepackage{textcomp}
\usepackage{stfloats}
\usepackage{url}
\usepackage{verbatim}
\usepackage{graphicx}
\usepackage{cite}
\usepackage{multirow}
\usepackage{tabularx}

\usepackage{orcidlink}
\usepackage{booktabs}
\usepackage{amssymb}
\usepackage{hyperref}
\usepackage{balance}
\begin{document}

\title{Temporal-Aware Spiking Transformer Hashing Based on 3D-DWT}

\author{Zihao Mei\(^{\orcidlink{0009-0003-0856-429X}}\), 
Jianhao Li\(^{\orcidlink{0009-0002-1628-1559}}\), 
Bolin Zhang\(^{\orcidlink{0000-0002-6133-9564}}\), 
Chong Wang\(^{\orcidlink{0000-0001-6016-6545}}\),
Lijun Guo\(^{\orcidlink{0000-0002-6133-9564}}\), 
Guoqi Li\(^{\orcidlink{0000-0002-8994-431X}}\),~\IEEEmembership{Member,~IEEE}, \\ and
Jiangbo Qian\(^{\orcidlink{0000-0003-4245-3246}}\),~\IEEEmembership{Member,~IEEE}
% <-this % stops a space
\thanks{Z. Mei, J. Li, B. Zhang, C. Wang, L. Guo and J. Qian are with the Faculty of Electrical Engineering and Computer Science, Ningbo University, Ningbo 315211, China (e-mail: 2211100269@nbu.edu.cn).}
\thanks{Guoqi Li is with the Institute of Automation, Chinese Academy of Sciences, and Key Laboratory of Brain Cognition and Brain-inspired Intelligence Technology, Beijing 100045, China.}
\thanks{Corresponding author: Jiangbo Qian.(e-mail: qianjiangbo@nbu.edu.cn)}
}

\markboth{Journal of \LaTeX\ Class Files,~Vol.~14, No.~8, August~2021}%
{Shell \MakeLowercase{\textit{et al.}}: A Sample Article Using IEEEtran.cls for IEEE Journals}


\maketitle

\begin{abstract}
  With the rapid growth of dynamic vision sensor (DVS) data, constructing a low-energy, efficient data retrieval system has become an urgent task. Hash learning is one of the most important retrieval technologies which can keep the distance between hash codes consistent with the distance between DVS data. As spiking neural networks (SNNs) can encode information through spikes, they demonstrate great potential in promoting energy efficiency. Based on the binary characteristics of SNNs, we first propose a novel supervised hashing method named Spikinghash with a hierarchical lightweight structure. Spiking WaveMixer (SWM) is deployed in shallow layers, utilizing a multilevel 3D discrete wavelet transform (3D-DWT) to decouple spatiotemporal features into various low-frequency and high-frequency components, and then employing efficient spectral feature fusion. SWM can effectively capture the temporal dependencies and local spatial features. Spiking Self-Attention (SSA) is deployed in deeper layers to further extract global spatiotemporal information. We also design a hash layer utilizing binary characteristic of SNNs, which integrates information over multiple time steps to generate final hash codes. Furthermore, we propose a new dynamic soft similarity loss for SNNs, which utilizes membrane potentials to construct a learnable similarity matrix as soft labels to fully capture the similarity differences between classes and compensate information loss in SNNs, thereby improving retrieval performance. Experiments on multiple datasets demonstrate that Spikinghash can achieve state-of-the-art results with low energy consumption and fewer parameters.
\end{abstract}

\begin{IEEEkeywords}
Spiking neural network, deep hashing, retrieval, neuromorphic computing.
\end{IEEEkeywords}

\section{Introduction}
\IEEEPARstart{D}{ynamic} vision sensor (DVS) is a novel form of visual data that encodes time, position, and polarity of each pixel's brightness change into event streams with microsecond-level temporal resolution, demonstrating significant advantages in many specific visual scenarios \cite{gallego2020event, amir2017low, rebecq2019high}. With the rapid growth of DVS data, retrieval tasks face severe challenges in terms of computational costs and energy consumption, and constructing a low-energy, efficient retrieval network has become an urgent task.

\begin{figure}[tb] \centering
  \includegraphics[width=0.48\textwidth,height=0.11\textwidth]{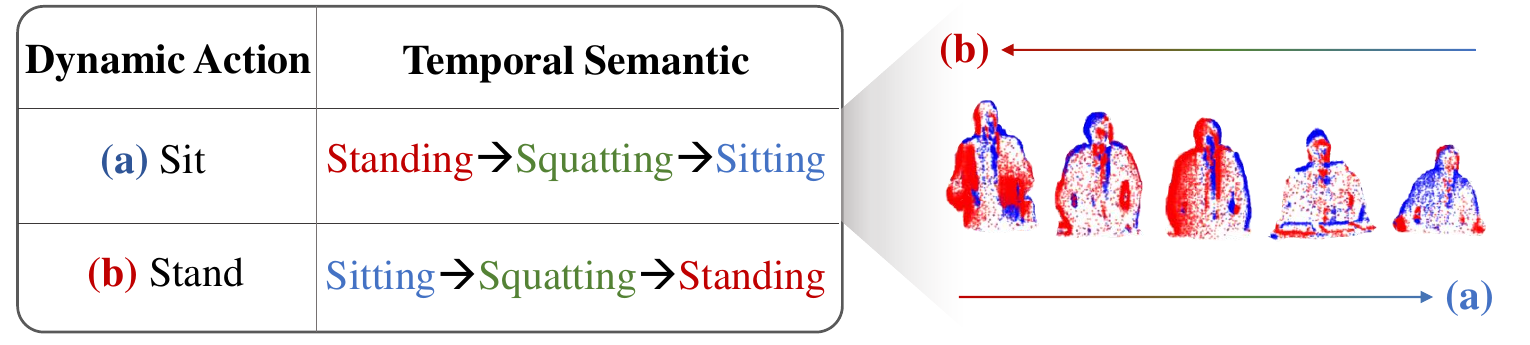}
  \caption{In HMDB51-DVS data, the actions ``sit" and ``stand" contain symmetric semantic information at different time steps. Distinguishing these actions requires recognizing the temporal order and dependencies across multiple time steps.} \label{fig:1}
  \vspace{-5pt}
\end{figure}

As the third generation of neural networks, spiking neural networks (SNNs) encode information through binary spikes, and have demonstrated significant advantages in terms of low energy consumption \cite{maass1997networks, pei2019towards}. SNNs show great potential for enhancing energy efficiency in large-scale DVS video and image retrievals. However, the current SNNs \cite{zhou2022spikformer, yao2024spike, zhou2023spikingformer, yao2024spike2} face some key challenges: (1) Although there is an urgent demand for efficient retrieval networks for large scale DVS data, no hashing methods have been proposed in SNNs. (2) Most existing methods design modules that consider only spatial information at a single time step, failing to capture temporal dependencies effectively across multiple time steps. As illustrated in Fig.~\ref{fig:1}, the actions ``sit" and ``stand" in HMDB51-DVS \cite{bi2020graph} contain symmetric semantic information at different time steps. Distinguishing these actions requires considering the temporal dependencies. However, models \cite{wang2023spatial, yao2023attention, yao2021temporal, zhu2024tcja, qiu2024gated} that consider the time steps, merely apply a simple weighting to each time step, or introduce computationally expensive modules, thereby failing to adequately capture temporal dependencies, increasing network overhead, and consequently struggle to distinguish between these two symmetric actions. (3) Current SNN-Transformers generally have large number of parameters. This results in substantial computational resource demands, hindering their deployment in resource-constrained environments. (4) Local spatial features are crucial for visual tasks. However, existing SNN-Transformers encounter constraints in their capacity to extract local spatial features due to the poor capabilities of the shallow convolutional networks executed before the Transformer encoders, and the limitations of the Spiking Self-Attention mechanisms \cite{shi2024spikingresformer}. These challenges hinder the current SNNs from achieving better performances in retrieval tasks.

To solve the above problems, we first propose a novel hashing method named Spikinghash based on the unique binary characteristics of SNNs. To effectively capture the temporal dependencies and local spatial information, we design an efficient wavelet feature fusion module named Spiking WaveMixer (SWM) based on multilevel 3D discrete wavelet transform (3D-DWT) \cite{weeks2002three}. SWM is deployed in shallow layers, utilizing multilevel 3D-DWT to decouple features into various low- and high-frequency components, and employing efficient spectral feature fusion. Spiking Self-Attention (SSA) is deployed in deeper layers to further extract global spatiotemporal information. The SSA module effectively integrates the fine-grained features learned by the SWM module, thereby achieving more comprehensive feature learning. Finally, we design a hash layer utilizing spiking neurons, which integrates information over multiple time steps to generate the final hash codes, eliminating the need for additional quantization modules. Here, we also propose a new dynamic soft similarity loss for SNNs. Spiking neurons convert floating-point membrane potentials into binary spikes, leading to information loss and consequently limiting the performance of SNNs. As membrane potentials contain richer feature information, we construct a learnable similarity matrix as soft labels based on membrane potentials. The similarity matrix is continuously updated and optimized during training, capturing more complex inter-class relationships and allowing valuable information to be retained through different rounds of feature learning. Compared to traditional hard similarity loss, this method more accurately reflects subtle similarity differences between classes, compensating the information loss in SNNs and thereby improving the retrieval performance. Experiments on multiple datasets demonstrate that Spikinghash can achieve state-of-the-art results with significant advantages in terms of low energy and parameter consumption. The main contributions of this paper can be summarized as follows:
\begin{itemize}
  \item We propose an efficient wavelet feature fusion module named Spiking WaveMixer (SWM). SWM uses a multilevel 3D-DWT to decouple features into various low-frequency and high-frequency components, effectively fusing these spectral features. This module significantly enhances the temporal dependencies and local spatial information extraction capabilities.
  \item Based on the binary characteristics of SNNs, we first propose a novel hashing method named Spikinghash. By embedding SWM in shallow layers and SSA in deep layers, Spikinghash fully integrates the strengths of extracting the local and global spatiotemporal features. This design effectively enhances performance while reducing the number of parameters.
  \item We propose a new dynamic soft similarity loss for SNNs that utilizes membrane potentials to construct a learnable similarity matrix as soft labels, capturing subtle inter-class similarity differences and reducing information loss in SNNs, thereby improving the retrieval performance.
  \item Experiments on multiple DVS video and image datasets show that Spikinghash can outperform state-of-the-art SNNs on both classification and retrieval tasks with significantly reduced parameters.
\end{itemize}

The rest of this paper is structured as follows: Section~\ref{sec:related_works} provides a brief review of related work. Section~\ref{sec:Method} details our proposed Spikinghash method. Section~\ref{sec:exp} demonstrates the effectiveness of Spikinghash through extensive experiments. Section~\ref{sec:conclusion} concludes our work.
\section{Related Work}
\label{sec:related_works}
\subsection{Spiking Neural Networks}%

Recently, integrating Transformers with SNNs to increase performance has emerged as a new research direction. Spikformer \cite{zhou2022spikformer} and Spikingformer \cite{zhou2023spikingformer} incorporate a self-attention mechanism specifically for SNNs, using spiking forms of \(\boldsymbol{Q}\), \(\boldsymbol{K}\), and \(\boldsymbol{V}\) to compute the attention matrix. Spike-driven Transformer \cite{yao2024spike} uses a unique linear attention, where matrix multiplication in self-attention is converted into Hadamard product-based masking operations, reducing energy consumption and simplifying the model. SpikingResformer \cite{shi2024spikingresformer} indicates that current SNN-Transformers employ shallow convolutional networks before the Transformer encoder to extract local features and reduce spatial dimensions. However, this shallow network has limited local feature extraction capabilities. Consequently, these models encounter constraints in their ability to extract local spatial features. Although these works have improved model performance by introducing Transformers to SNNs, they lack consideration of the timestep dimension of SNNs, resulting in suboptimal performances, especially for event-based datasets with rich motion information.

Most of current SNNs lack consideration for temporal dependencies. Some recent methods enhance focus on the temporal dimension typically involves integrating attention mechanisms to assess the importance of different time steps. STS-Transformer \cite{wang2023spatial} introduces the spatial-temporal self-attention mechanism and spatial-temporal relative position bias (STRPB), preserving the asynchronous nature of SNNs. Some works \cite{yao2021temporal, zhu2024tcja, qiu2024gated, yao2023attention} employ temporal attention to learn frame-level event stream representations, which filter irrelevant frames during inference. While these methods improve performance by introducing extra modules that focus on the temporal dimension, they also increase the number of parameters.

\subsection{Frequency-domain Deep Learning}%
Introducing frequency domain transformation into deep learning presents a novel perspective that effectively reduces the computational complexity and parameters while maintaining high model performance. In ANNs, various methods have been explored. Guibas et al. \cite{guibas2021efficient} proposed a model for feature fusion in the Fourier domain, enabling efficient global convolution in the frequency domain. SVT \cite{patro2024scattering} incorporates a dual-tree complex wavelet transform and separately processes low-frequency and high-frequency components. WaveMLP \cite{tang2022image} represents each token as a wave function with amplitude and phase parts, and it is capable of modeling different content in various input images. Additionally, WaveViT \cite{yao2022wave} can reduce the complexity of self-attention by introducing a wavelet transform for lossless downsampling.

\subsection{Learning-Based Hashing}%
Compared with traditional methods, deep hashing maps data into Hamming space, enabling faster XOR operations for computing similarities and reducing storage requirements. The hash codes aim to preserve the similarity structure with the original data, ensuring that similar samples have small Hamming distances. Deep hashing has been widely applied in image retrieval. HashNet \cite{cao2017hashnet} uses a continuous scale strategy to optimize continuous hash codes into discrete ones, reducing quantization loss. DPN \cite{fan2020deep} introduces a differentiable polarization loss function to minimize Hamming distances. TransHash \cite{chen2022transhash} and HashFormer \cite{li2022hashformer} are the first hashing networks that employ the Transformer architecture. Deep hashing has also been applied in video retrieval. SNPH \cite{li2021structure} uses clustering for pseudo labels on the basis of neighborhood similarity, and reconstruction tasks, for feature learning. BTH \cite{l2021self} introduces BERT to reconstruct frame features, and employed clustering to create pseudo labels. CONMH \cite{wang2023contrastive} is a single-stage hashing method based on contrastive learning that maximizes the similarity of positive samples with debiased contrastive loss. We first extend deep hashing to SNNs and design a novel supervised hashing method based on the binary characteristics of SNNs.

\subsection{Summary}%
Artificial neural networks (ANNs) inevitably face high computational costs and high energy consumption when processing large volumes of DVS data; hence, constructing a low-energy, efficient retrieval network has become an urgent task. Spiking neural networks (SNNs), which encode information through spikes, have significant advantages in terms of low energy consumption. However, the performance of SNNs in retrieval tasks is limited by challenges such as large number of parameters, lack of temporal dependencies, and the loss of local spatial features. To overcome these challenges and optimize retrieval performance, we specifically introduce a multilevel 3D discrete wavelet transform (3D-DWT) into SNNs and design a novel supervised hashing method named Spikinghash with a hierarchical lightweight structure. This method not only effectively reduces the number of parameters but also enhances the extraction of temporal dependencies and local spatial features, thereby achieving an optimized balance between performance and efficiency.

\section{Method}
\label{sec:Method}
Spikinghash incorporates three modules: the Spiking Waveformer Block, the Spiking Transformer Block, and the spiking hash layer. We first outline the basic principles of spiking neurons, followed by a detailed introduction to the overall architecture of Spikinghash. Finally, we introduce the dynamic soft similarity loss function.

\subsection{Spiking Neuron Layer}%
Spiking neuron is the essential core component in SNNs. Nowadays, various spiking neuron models have been proposed, including leaky integer-and-fire (LIF) model \cite{teeter2018generalized}, Hodgkin-Huxley (H-H) model \cite{hodgkin1952quantitative}, and Izhikevich model \cite{izhikevich2003simple}. We uniformly use the LIF model in experiments. The LIF neuron model iterates over multiple time steps, with its internal operations divided into three processes. The formulas are as follows:
\begin{align}
  H[t] &= V[t-1] + \frac{1}{\gamma} \left(I[t] - (V[t-1] - V_{\text{reset}})\right), \label{eq:1} \tag{1}\\
  S[t] &= \text{Hea}(H[t] - v_{\text{th}}), \label{eq:2} \tag{2}\\
  V[t] &= V_{\text{reset}} S[t] + H[t](1 - S[t]). \label{eq:3}\tag{3}
\end{align}

Eq.~\eqref{eq:1} represents the charging process, where \(H[t]\) is the membrane potential generated by combining spatial input from the current time step \(I[t]\) with the reset membrane potential from the previous time step \(V[t-1]\). The membrane time constant \(\gamma\) attenuates the input. Eq.~\eqref{eq:2} represents the firing process, where \(\mathrm{Hea}(\cdot)\) denotes the Heaviside step function. When the membrane potential \(H[t]\) exceeds the threshold \(v_{\text{th}}\), \(S[t] = 1\), indicating that the spiking neuron fires a spike. Conversely, \(S[t] = 0\), indicating that no spike is fired. Eq.~\eqref{eq:3} represents the resetting process, where \(V[t]\) is the membrane potential after resetting at the current time step. If a spike is fired, \(V[t]\) is set to the reset potential \(V_{\text{reset}}\); otherwise, it retains the original membrane potential \(H[t]\).

\begin{figure*}[tb] \centering
  \includegraphics[width=0.9\textwidth,height=0.53\textwidth]{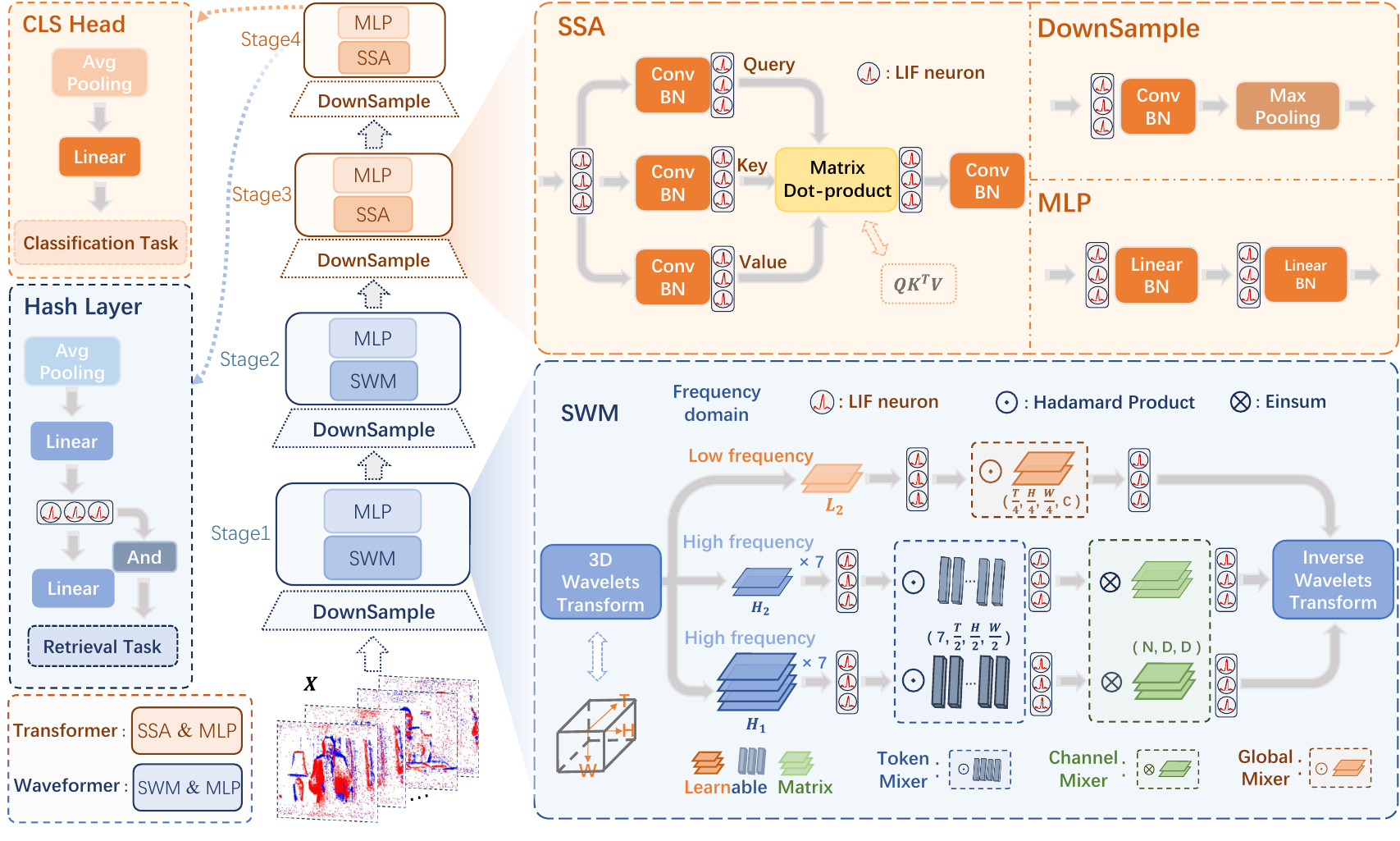}
  \caption{The overall architecture of Spikinghash. This hierarchical SNN-Transformer architecture includes a downsample layer before each stage. The first two stages deploy Spiking Waveformer Blocks (including Spiking WaveMixer (SWM) and MLP). The last two stages deploy Spiking Transformer Blocks (including Spiking Self-Attention (SSA) and MLP). The residual connections in the Spiking Waveformer Blocks and Spiking Transformer Blocks are omitted in the figure. Hash layer or classification head is connected depending on the downstream task.} \label{fig:2}
\end{figure*}
%%\vspace{-0.8em}
\subsection{Overall Architecture}%
The overall architecture of Spikinghash is shown in Fig.~\ref{fig:2}. The input data is represented as \( \boldsymbol{X} \). Spikinghash is structured into four stages, with a downsample layer applied before each stage. On the basis of the characteristics of each module, the first two stages deploy Spiking Waveformer Blocks (including Spiking WaveMixer (SWM) and multilayer perceptron (MLP)), whereas the last two stages deploy Spiking Transformer Blocks (including Spiking Self-Attention (SSA) and MLP). This hierarchical structure sequentially extracts features from the frequency domain to the spatiotemporal domain, leveraging the strengths of SWM in extracting the local spatial features and temporal dependencies, and SSA for its ability to extract global spatiotemporal information. After feature extraction through the hierarchical stages, the model ends with the spiking hash layer. To validate the efficiency of Spikinghash, we additionally included classification tasks, where the model ends with a classification layer. The details of the overall architecture of Spikinghash series are listed in Table~\ref{tab:1}. In the following sections, we introduce three modules in Spikinghash: Spiking Waveformer Block, Spiking Transformer Block, and Spiking Hash Layer.
%\vspace{-0.5em}
\subsection{Spiking Waveformer Block}%
To enhance the capture of temporal dependencies and local spatial features, the Spiking Waveformer Block is proposed, which consists of a Spiking WaveMixer (SWM) layer and an MLP layer, as shown in Fig.~\ref{fig:2}. The formulas are as follows:
\begin{align}
  \boldsymbol{X} &= \boldsymbol{X} + \mathrm{SWM}(\boldsymbol{X}), \label{eq:4} \tag{4}\\
  \mathrm{MLP}(\boldsymbol{X}) &= \mathrm{BN}(\mathrm{Linear}(\mathrm{SN}(\mathrm{BN}(\mathrm{Linear}(\mathrm{SN}(\boldsymbol{X})))))), \label{eq:5} \tag{5}\\
  \boldsymbol{X} &= \boldsymbol{X} + \mathrm{MLP}(\boldsymbol{X}), \label{eq:6} \tag{6}
\end{align}
where \(\mathrm{SN}(\cdot)\) represents the spiking neurons, \(\mathrm{BN}(\cdot)\) denotes the batch normalization layer, and \(\mathrm{Linear}(\cdot)\) represents the linear layer.

The Spiking Waveformer Block is deployed in shallow layers to extract temporal dependencies and better leverage its ability to extract local spatial features. Specifically, shallow layers retain more fine-grained spatial information, while deeper layers tend to abstract higher-level features. SWM applies multilevel DWT in the spatiotemporal dimension, reducing the spatial resolution by \(2^N\) times (where N is the number of wavelet transform levels) and extracting spectral features at multiple scales. Using SWM in shallow layers ensures that local spatial features are captured effectively before they are aggregated or lost in deeper layers. In the following sections, we outline the 3D-DWT in SWM and provide a detailed overview of SWM.
\subsubsection{3D Discrete Wavelet Transform}
3D discrete wavelet transform (3D-DWT) decomposes multidimensional data into low-frequency and high-frequency components at various scales and directions \cite{weeks2002three}. In this work, we introduce 3D-DWT into SNNs for the first time, aiming to capture the dynamic changes in spikes over timesteps and the spatial details. We employ 3D-DWT to decompose the input features \( \boldsymbol{X} \in \mathbb{R}^{T \times C \times H \times W} \). Initially, it is reshaped to \(\mathbb{R}^{C \times T \times H \times W}\), followed by DWT along the dimensions of the timestep (\( T \)), height (\( H \)), and width (\( W \)). To maintain the sparse additive property of SNNs, we convert the DWT iutputs of each dimension into spike features through spiking neurons \( \mathrm{SN}(\cdot) \). Its mathematical formulas are as follows:
\begin{align}
  \mathrm{DWT}_d\left(x[n]\right) &\left\{ 
    \begin{array}{ll}
      a[n]&= \sum_{k=0}^{k-1} x[2n-k]l[k] \\
      d[n]&= \sum_{k=0}^{k-1} x[2n-k]h[k]
    \end{array},
    \right. \label{eq:7}\tag{7}
\end{align}
%\vspace{-1.0em}
\begingroup
\small{
\begin{align}
  \mathrm{DWT}_{3D}(\boldsymbol{X}) &= \mathrm{DWT}_W(\mathrm{SN}(\mathrm{DWT}_H(\mathrm{SN}(\mathrm{DWT}_T(\mathrm{SN}(\boldsymbol{X})))))), \label{eq:8}\tag{8}
\end{align}
}
\endgroup
where \(\mathrm{DWT}_d(\cdot)\) denotes the application of the 1D-DWT along dimension \( d \), \( x[n] \) represents the data sequence of \(\boldsymbol{X} \) along dimension \( d \), \( l[\cdot] \) and \( h[\cdot] \) are the low-pass and high-pass filters, respectively; and \( k \) is the index of the filter coefficients. \( a[n] \) denotes the generated low-frequency component, whereas \( d[n] \) represents the high-frequency component. Considering the simplicity and linear computational complexity of a Haar wavelet transform, we utilize the Haar wavelet as the basis for the 3D-DWT. Specifically, the coefficients of \( l[\cdot] \) are \(\frac{1}{\sqrt{2}}[1,1]\), and the coefficients of \( h[\cdot] \) are \(\frac{1}{\sqrt{2}}[1,-1]\).

\begin{figure}[tb]
  \centering
  \includegraphics[width=0.42\textwidth,height=0.35\textwidth]{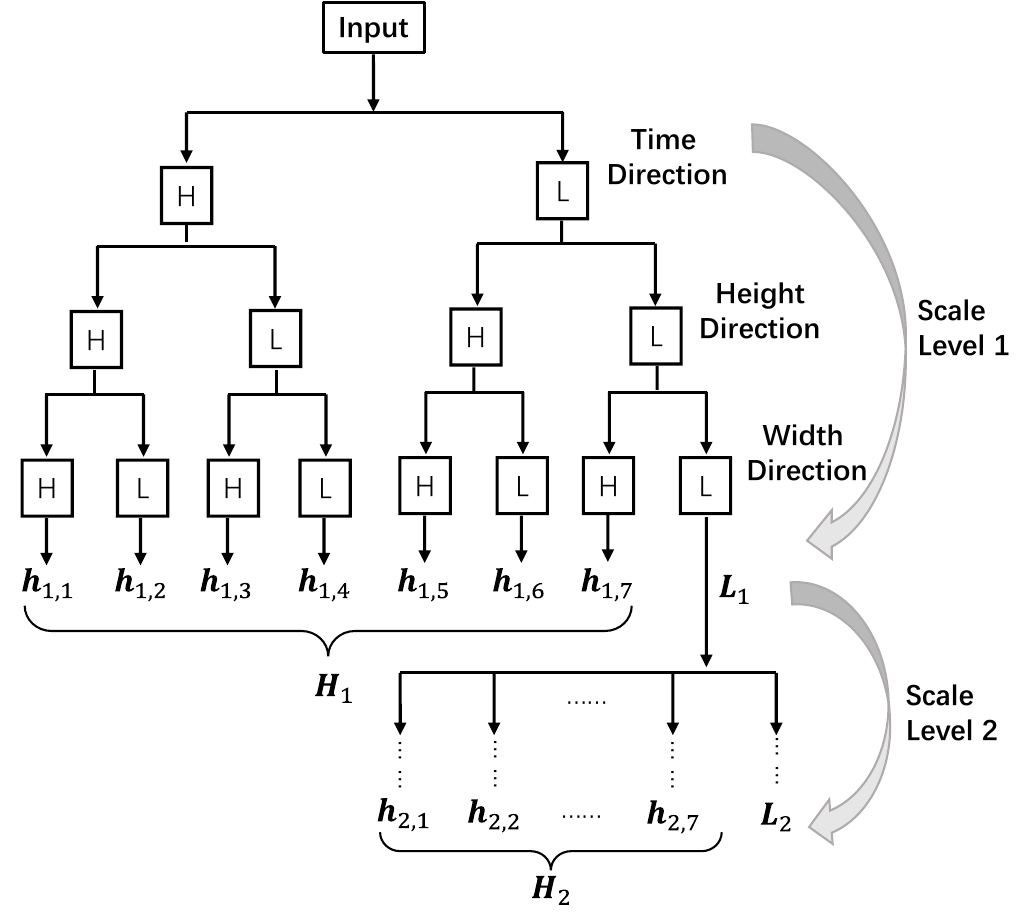}
  \caption{The iterative multilevel decomposition process of 3D-DWT.}\label{fig:3}
\end{figure}

After completing the first-level 3D-DWT, we obtain seven different directional high-frequency components \( \boldsymbol{H}_1 = \{ \boldsymbol{h}_{1,1}, \boldsymbol{h}_{1,2}, \ldots, \boldsymbol{h}_{1,7} \} \in \mathbb{R}^{7 \times C \times \frac{T}{2} \times \frac{H}{2} \times \frac{W}{2}} \), and one low-frequency component \( \boldsymbol{L}_1 \in \mathbb{R}^{C \times \frac{T}{2} \times \frac{H}{2} \times \frac{W}{2}} \). To extract finer local spatial features, we employ a multilevel 3D-DWT. As shown in Fig.~\ref{fig:3}, multilevel 3D-DWT is achieved by recursively decomposing the low-frequency component. Upon further decomposition of the low-frequency component \( \boldsymbol{L}_1 \), we obtain seven high-frequency components \( \boldsymbol{H}_2 = \{ \boldsymbol{h}_{2,1}, \boldsymbol{h}_{2,2}, \ldots, \boldsymbol{h}_{2,7} \} \in \mathbb{R}^{7 \times C \times \frac{T}{4} \times \frac{H}{4} \times \frac{W}{4}} \), and one low-frequency component  \( \boldsymbol{L}_2 \in \mathbb{R}^{C \times \frac{T}{4} \times \frac{H}{4} \times \frac{W}{4}} \) for the second-level 3D-DWT. In our experiments, we use two levels of 3D-DWT by default. After two levels of the 3D-DWT, the features are decoupled into high-frequency components at different scales and directions, \( \boldsymbol{H}_1 \) and \( \boldsymbol{H}_2 \), and one low-frequency component \( \boldsymbol{L}_2 \). These components are input into the subsequent module for spectral feature fusion.
\subsubsection{Spiking WaveMixer}
We use a novel method named Spiking WaveMixer (SWM, Fig.~\ref{fig:2}) to fuse and transform these spectral features on the basis of the characteristics of the high-frequency and low-frequency components.
The low-frequency component \( \boldsymbol{L}_2 \) reflects smooth variations over larger regions, representing global spatiotemporal information and having smaller dimensions. We utilize a global mixer (Hadamard product) to transform the features of the low-frequency component, aiming to fully exploit the global spatiotemporal information in \( \boldsymbol{L}_2 \).

Unlike the low-frequency component, high-frequency components involve more complex spatial relationships and dynamic information, incorporating elements across multiple scales and directions. The high-frequency components represent local spatial features such as the edges and textures in the spatial information, whereas in the temporal information, they reflect parts with significant changes, typically foreground information that needs attention. On the other hand, DVS data are derived from changes in brightness, such as object motion or variations in lighting, which represent high-frequency information. On the basis of the characteristics of high-frequency components and their importance to DVS data and SNNs, we adopt a more detailed stepwise feature fusion strategy to effectively capture complex spatial information and temporal dynamics. Specifically, we divide the feature transformation into two steps: first, a token mixer is performed on the high-frequency components to update features across multiple dimensions, including different scales, directions, and spatial-temporal dimensions, and then a channel mixer is used to fuse features along the channel dimension.

In the global mixer, \( \boldsymbol{L}_2 \) is initially transformed into spike features through spiking neurons \( \mathrm{SN}(\cdot) \), followed by Hadamard product (\(\circ)\) with a learnable parameter matrix \( \boldsymbol{W}_1 \in \mathbb{R}^{C \times \frac{T}{4} \times \frac{H}{4} \times \frac{W}{4}} \). The formula is as follows:
\begin{align}
  \boldsymbol{L}^\prime &= \mathrm{Global\ Mixer}(\boldsymbol{L}_2, \boldsymbol{W}_1) = \mathrm{SN}(\boldsymbol{L}_2) \circ \boldsymbol{W}_1. \tag{9}
\end{align}

In the token mixer, the high-frequency components, \( \boldsymbol{H}_1 \) and \( \boldsymbol{H}_2 \), are initially transformed into spike features through spiking neurons \( \mathrm{SN}(\cdot) \), followed by Hadamard product (\(\circ)\) with learnable parameter matrices \( \boldsymbol{W}_2 \in \mathbb{R}^{7 \times \frac{T}{2} \times \frac{H}{2} \times \frac{W}{2}} \) and \( \boldsymbol{W}_3 \in \mathbb{R}^{7 \times \frac{T}{4} \times \frac{H}{4} \times \frac{W}{4}} \), respectively. The formula is as follows:
\begin{align}
  \mathrm{Token\ Mixer} \left\{ 
    \begin{array}{ll}
      \boldsymbol{H}_1' &= \mathrm{SN}(\boldsymbol{H}_1) \circ \boldsymbol{W}_2 \\
      \boldsymbol{H}_2' &= \mathrm{SN}(\boldsymbol{H}_2) \circ \boldsymbol{W}_3
    \end{array}.
  \right. \tag{10}
\end{align}
We employ the Hadamard product in both the token mixer and global mixer for its simplicity and efficiency. In both the token mixer and the global mixer, features at different time steps are adaptively assigned distinct weights, allowing the features at each time step to be emphasized or attenuated on the basis of their importance in the sequence.

After the token mixer, we perform the channel mixer to further integrate features along the channel dimension. We utilize the Einstein summation convention \cite{patro2024scattering} for the channel mixer because of its flexibility and simplicity: First, high-frequency components are divided into several groups along the channel dimension, similar to the multihead attention mechanism, to reduce the number of parameters. Specifically, they are reshaped into \( \boldsymbol{H}_1' \in \mathbb{R}^{7 \times \frac{T}{2} \times \frac{H}{2} \times \frac{W}{2} \times N \times D} \) and \( \boldsymbol{H}_2' \in \mathbb{R}^{7 \times \frac{T}{4} \times \frac{H}{4} \times \frac{W}{4} \times N \times D} \), where \( N \times D = C \), \( N \) is the number of groups, and \( D \) is the channel dimension of each group. We then utilize learnable parameter matrices \( \boldsymbol{W}_4, \boldsymbol{W}_5 \in \mathbb{R}^{N \times D \times K} \) to fuse the high-frequency components \(\boldsymbol{H}_1' \) and \( \boldsymbol{H}_2' \) through the Einstein summation convention. The formulas are as follows:
\begin{align}
  \mathrm{Channel\ Mixer} \left\{ 
    \begin{array}{ll}
      \boldsymbol{H}_1'' &= \mathrm{Einsum}(\mathrm{SN}(\boldsymbol{H}_1'), \boldsymbol{W}_4) \\
      \boldsymbol{H}_2'' &= \mathrm{Einsum}(\mathrm{SN}(\boldsymbol{H}_2'), \boldsymbol{W}_5)
    \end{array},
  \right. \tag{11}
\end{align}

\begin{align}
  \mathrm{Einsum}(\boldsymbol{I}, \boldsymbol{W}) &= \sum_{d=0}^{D-1} (\boldsymbol{I}_d \circ \boldsymbol{W}_d). \tag{12}
\end{align}

Finally, we apply the 3D inverse discrete wavelet transform (3D-iDWT) to the enhanced low-frequency component \(\boldsymbol{L}^\prime \) and high-frequency components \( \boldsymbol{H}_1^{\prime\prime} \) and \( \boldsymbol{H}_2^{\prime\prime} \), reverting the features from the frequency domain back to the spatiotemporal domain.

\subsection{Spiking Transformer Block}%
After the Spiking Waveformer Block, the deeper layers in the network possess richer semantic information and larger receptive fields, facilitating the extraction of more abstract global spatiotemporal features. Therefore, in the last two stages of the model, we employ the Spiking Transformer Block \cite{zhou2023spikingformer}, which consists of two modules: Spiking Self-Attention (SSA) and MLP (Fig.~\ref{fig:2}). The formulas are as follows:
\begin{align}
  &\boldsymbol{Q}, \boldsymbol{K}, \boldsymbol{V} = \mathrm{SN}(\mathrm{BN}(\mathrm{Conv}(\boldsymbol{X}))), \tag{13} \\
  &\mathrm{SA}(\boldsymbol{Q}, \boldsymbol{K}, \boldsymbol{V}) = \mathrm{SN}(\boldsymbol{Q}\boldsymbol{K}^T \boldsymbol{V} \ast s), \tag{14} \\
  &\boldsymbol{X} = \boldsymbol{X} + \mathrm{BN}(\mathrm{Conv}(\mathrm{SA}(\boldsymbol{Q}, \boldsymbol{K}, \boldsymbol{V}))), \tag{15} \\
  &\boldsymbol{X} = \boldsymbol{X} + \mathrm{MLP}(\boldsymbol{X}), \tag{16}
\end{align}
where \(s\) represents the scaling factor, \(\mathrm{Conv}(\cdot)\) denotes a convolutional layer with a kernel size of \(1 \times 1\), and \(\mathrm{MLP}(\cdot)\) follows the same formula as Eq.~\ref{eq:5}.

Utilizing the SSA in deeper layers allows for further attention to the global relationships among different parts of the visual features, thereby achieving more refined and comprehensive feature representations. Compared with traditional single attention mechanisms, the hierarchical architecture, which combines global self-attention with SWM's ability to extract local spatial features and temporal dependencies, enables the model to preserve crucial local details while capturing the global spatiotemporal information, thereby enhancing the model's expressive capacity.

\subsection{Spiking Hash Layer}%
\subsubsection{Generating Hash Codes}
We first extend deep hashing to SNNs to construct an efficient retrieval network with low energy consumption and storage requirements. After feature extraction via Spikinghash and global average pooling, we can obtain a feature vector represented as \( \boldsymbol{f}_i \in \mathbb{R}^{T \times C} \). For traditional hash layers, the channel dimension \( C \) is mapped to the length \( L \) of the hash codes through a linear layer \(\mathrm{Linear}(\cdot)\). The formula for obtaining the hash code \(\boldsymbol{b}_i\) is as follows:
\begin{align}
  \boldsymbol{b}_i = \mathrm{sgn}(\tanh(\mathrm{Linear}(\boldsymbol{f}_i))), \tag{17}
\end{align}
\noindent
where \(\tanh(\cdot)\) is a hyperbolic tangent function, \(\mathrm{sgn}(\cdot)\) is a sign function, and \( \boldsymbol{b}_i \) is the binary hash code.

SNNs encode information through spikes, which naturally aligns with the binary format of the hash codes. Based on the binary characteristics of SNNs, we design a novel hashing method. Specifically, we directly use the spikes as hash codes, which eliminates the need for an additional quantization layer. We first map the feature dimension from \( C \) to the hash code length \( L \) through a linear layer. The features are then input into spiking neurons to obtain spike features across multiple time steps. To reduce the hash code dimensionality across multiple time steps and leverage the temporal information of SNNs, we apply a bitwise AND operation on the spikes across multiple time steps, yielding the final hash code. The formula is as follows:
\begin{align}
  \boldsymbol{b}_i = \mathrm{AND}\left(\mathrm{SN}\left(\mathrm{Linear}(\boldsymbol{f}_i)\right)\right), \tag{18}
\end{align}
\noindent
where \(\mathrm{AND}(\cdot)\) represents the bitwise AND operation along the time steps, and \(\mathrm{SN}(\cdot)\) denotes spiking neurons. After obtaining the hash codes, we optimize them with a new loss function for SNNs.

\subsection{Loss function}
\subsubsection{Dynamic soft similarity loss}
Hash codes typically need to be optimized by loss functions to preserve the similarity among samples. After spiking hash layer, the hash codes are denoted as \( \boldsymbol{B} = \{ \boldsymbol{b}_i \}_{i=1}^N \in \mathbb{R}^{N \times L} \), where \( N \) is the number of samples and \( L \) is the length of the hash codes. The corresponding labels are denoted as \( \boldsymbol{Y} = \{ \boldsymbol{y}_i \}_{i=1}^N \in \mathbb{R}^{N} \). Traditional deep hashing methods typically generate a binary similarity matrix \(\boldsymbol{M}_{\text{hard}} = \{m_{ij}\}_{i,j=1}^N \in \{0,1\}^{N \times N}\) from the label \( \boldsymbol{Y} \) to construct the similarity loss. If samples \( \boldsymbol{b}_i \) and \( \boldsymbol{b}_j \) belong to the same category, then \( m_{ij} = 1 \); otherwise, \( m_{ij} = 0 \). However, this predefined binary similarity matrix \( \boldsymbol{M_\text{hard}} \) overlooks the potential similarity differences between classes. For example, in UCF101-DVS, ``kayaking" should be considered more similar to ``rafting" than ``haircut", while they are treated as equally different classes in traditional similarity loss.

Spiking neurons convert floating-point membrane potentials into binary spikes, which leads to information loss and limits the performance of SNNs. As membrane potentials contain richer information compared to binary spikes, we constructed a learnable similarity matrix \(\boldsymbol{S}_{\text{soft}} = \{s_{ij}\}_{i,j=1}^C \in \mathbb{R}^{C \times C}\) as soft labels to capture more complex inter-class relationships and reduce the information loss in SNNs, where \( C \) is the number of classes. 

We divide the total training epochs into different rounds. In each round, the similarity matrix \(\boldsymbol{S}_{\text{soft}}\) is updated based on the membrane potentials of the current round and used for supervision in the next round. Based on the classification results, the similarity $r$ between the correctly classified samples is computed to update the similarity matrix \(\boldsymbol{S}_{\text{soft}}\). The formulas are as follows:
\begin{equation}
  \left\{
  \begin{aligned}
  r_{ij} &= \frac{{\boldsymbol{h}}_i \cdot {\boldsymbol{h}}_j^T}{\|{\boldsymbol{h}}_i\| \cdot \|{\boldsymbol{h}}_j\|}, \\
  s_{y_iy_j} &= s_{y_iy_j} + r_{ij}, \\
  k_{y_iy_j} &= k_{y_iy_j} + 1.
  \end{aligned}
  \right.
  \tag{19}
\end{equation}
Here, \(\boldsymbol{H} = \left\{ \boldsymbol{h}_i \right\}_{i=1}^N \in \mathbb{R}^{N \times L}\) denotes the membrane potentials (averaged over time steps) retained by spiking neurons in the hash layer. \(\boldsymbol{K} = \left\{ k_{ij} \right\}_{i,j=1}^C \in \mathbb{R}^{C \times C}\) is an accumulated value to normalize \(\boldsymbol{S}_{\text{soft}}\), where \(\boldsymbol{S}_\text{soft} = \frac{\boldsymbol{S}_\text{soft}}{\boldsymbol{K}}\), at the end of each round. Additionally, to reduce noise interference, we filter similarity matrix \(\boldsymbol{S}_{\text{soft}}\) to obtain the final similarity matrix. The formula is as follows:
\begin{equation}
  s_{ij} = 
  \begin{cases}
  s_{ij}, & \text{if } s_{ij} > \tau \\
  0, & \text{otherwise}
  \end{cases},
  \tag{20}
\end{equation}
where \(\tau\) is the filtering threshold. 

During training, we gradually transition from a hard similarity matrix to a soft similarity matrix to prevent interference from similar classes on the feature learning of the same classes. The formula is as follows:
\begin{equation}
  \boldsymbol{S} = \lambda \cdot \boldsymbol{S_{\text{hard}}} + (1 - \lambda) \cdot \boldsymbol{S_{\text{soft}}},\tag{21}
\end{equation}
where $\lambda$ gradually decreases from 1 to 0 through the training process. We use \(\boldsymbol{S} = \left\{ s_{ij}' \right\}_{i,j=1}^C \in \mathbb{R}^{C \times C}\) as soft labels to obtain the corresponding similarity matrix \(\boldsymbol{M_{\text{soft}}} = \left\{ m_{ij}' \right\}_{i,j=1}^N \in \mathbb{R}^{N \times N}, \)
where $m_{ij}' = s_{y_iy_j}'$. The similarity matrix obtained by this method can more accurately reflect inter-class similarity differences, thereby improving retrieval performance. After obtaining \(\boldsymbol{M_{\text{soft}}}\), which represents the soft similarity relationships between sample pairs, the negative log-likelihood is used to measure the pairwise similarity loss, defined as follows:
\begin{align}
  L_{\mathrm{s}} &= -\sum_{i=1}^{N} \sum_{j=1}^{N} \log\left(p\left(m_{ij}' \mid \boldsymbol{b}_i, \boldsymbol{b}_j \right)\right) \nonumber \\
  & = -\sum_{i=1}^{N} \sum_{j=1}^{N} \Big( m_{ij}' \log\left(\sigma\left(\Omega_{ij}\right)\right) \nonumber\\
  & \quad + (1 - m_{ij}') \log\left(1 - \sigma\left(\Omega_{ij}\right)\right) \Big) \nonumber\\
  &= \sum_{i=1}^{N} \sum_{j=1}^{N} \left( \log\left(1+e^{\frac{1}{2}\boldsymbol{b}_i^T \boldsymbol{b}_j}\right) - \frac{1}{2} m_{ij}' \boldsymbol{b}_i^T \boldsymbol{b}_j \right), \tag{22}
\end{align}
where the inner product \(\Omega_{ij} = \frac{1}{2}\langle \boldsymbol{b}_i, \boldsymbol{b}_j \rangle \). The sigmoid function \(\sigma(\cdot)\) is used to convert the Hamming distance into a similarity score.
\subsubsection{Overall loss}
Additionally, to fully utilize the label information, we feed the hash codes through a classification layer to obtain classification results and incorporate the classification loss to jointly optimize the hash codes. The final loss is shown in the following formulas:
\begin{align}
  L_{\mathrm{cls}} &= -\sum_{t=1}^{C} y_t \log(p_t), \tag{23} \\
  L &= \alpha \cdot L_{\mathrm{s}} + \beta \cdot L_{\mathrm{cls}}, \tag{24}
\end{align}
where \(L_{\mathrm{cls}}\) denotes the classification loss, and where \(p_t\) represents the predicted probability for the \(t\)-th class. The parameters \(\alpha\) and \(\beta\) are used to balance the two losses. We backpropagate the gradients of the loss \(L\) via backpropagation through time (BPTT), as described in \cite{wu2018spatio}.
\section{Experimental Evaluation}
\label{sec:exp}
We evaluate Spikinghash on both DVS datasets of HARDVS \cite{wang2024hardvs}, CIFAR10-DVS \cite{li2017cifar10}, DVS128 Gesture \cite{amir2017low}, UCF101-DVS \cite{bi2020graph}, and HMDB51-DVS \cite{bi2020graph}, and static datasets of Tiny-ImageNet \cite{deng2009imagenet} and CIFAR10/CIFAR100 \cite{krizhevsky2009learning}. We implement Spikinghash via SpikingJelly \cite{fang2023spikingjelly} and trained the model from scratch. The details of the overall architecture of the Spikinghash series are listed in Table~\ref{tab:1}. In Section~\ref{sec:1}, we evaluate the energy consumption of Spikinghash. In Section~\ref{sec:2}, we compare the performance on hashing retrieval tasks with that of the current deep hashing methods. In Section~\ref{sec:3} and Section~\ref{sec:4}, we compare the performance of Spikinghash with that of the current state-of-the-art (SOTA) SNNs on the action recognition and image classification tasks. Finally, in Section~\ref{sec:5}, we conduct ablation experiments.
\begin{table*}[ht]
  \begin{center}
  \caption{Architectures of Spikinghash series.}
  \label{tab:1}
  \resizebox{0.75\textwidth}{!}{
  \begin{tabular}{lcccccc}
    \toprule
    \textbf{Stage} & \textbf{Layer Name} & \textbf{Spikinghash-Ti} & \textbf{Spikinghash-S} & \textbf{Spikinghash-M} & \textbf{Spikinghash-L} & \textbf{Spikinghash-XL} \\
    \midrule
    Stem & Stem & \multicolumn{5}{c}{[Conv 3×3, stride = 1]} \\
    \midrule
    \multirow{4}{*}{Stage 1} & DownSample & [SN; Conv; MP] & [SN; Conv] & [SN; Conv] & [SN; Conv; MP] & [SN; Conv; MP] \\
    \cline{2-7}
    \addlinespace
    & SWM & \multicolumn{1}{c}{\multirow{2}{*}{-}} & \multicolumn{1}{c}{\multirow{2}{*}{$\left[\begin{array}{c} D_1=96 \\ N_1=16N \end{array}\right]$}} & \multicolumn{1}{c}{\multirow{2}{*}{$\left[\begin{array}{c} D_1=96 \\ N_1=16N \end{array}\right]$}} & \multicolumn{1}{c}{\multirow{2}{*}{$\left[\begin{array}{c} D_1=128 \\ N_1=64N \end{array}\right]$}} & \multicolumn{1}{c}{\multirow{2}{*}{$\left[\begin{array}{c} D_1=128 \\ N_1=64N \end{array}\right]$}}\\
    & MLP & \multicolumn{5}{c}{}\\ 
    \addlinespace
    \midrule
    \multirow{4}{*}{Stage 2} & DownSample & [SN; Conv; MP] & [SN; Conv] & [SN; Conv] & [SN; Conv; MP] & [SN; Conv; MP] \\
    \cline{2-7}
    \addlinespace
    & SWM & \multicolumn{1}{c}{\multirow{2}{*}{$\left[\begin{array}{c} D_2=128 \\ N_2=16N \end{array}\right]$}} & \multicolumn{1}{c}{\multirow{2}{*}{$\left[\begin{array}{c} D_2=192 \\ N_2=16N \end{array}\right]$}} & \multicolumn{1}{c}{\multirow{2}{*}{$\left[\begin{array}{c} D_2=192 \\ N_2=16N \end{array}\right]$}} & \multicolumn{1}{c}{\multirow{2}{*}{$\left[\begin{array}{c} D_2=256 \\ N_2=16N \end{array}\right]$}} & \multicolumn{1}{c}{\multirow{2}{*}{$\left[\begin{array}{c} D_2=256 \\ N_2=16N \end{array}\right]$}}\\
    & MLP & \multicolumn{5}{c}{} \\ 
    \addlinespace
    \midrule
    \multirow{4}{*}{Stage 3} & DownSample & \multicolumn{5}{c}{[SN]; [Conv 3×3, stride = 1]; [Maxpooling 3×3, stride = 2]}\\
    \cline{2-7}
    \addlinespace
    & SWM & \multicolumn{1}{c}{\multirow{2}{*}{-}} & \multicolumn{1}{c}{\multirow{2}{*}{$\left[\begin{array}{c} D_3=192 \\ N_3=4N \end{array}\right]$}} & \multicolumn{1}{c}{\multirow{2}{*}{$\left[\begin{array}{c} D_3=384 \\ N_3=4N \end{array}\right]$}} & \multicolumn{1}{c}{\multirow{2}{*}{$\left[\begin{array}{c} D_3=384 \\ N_3=4N \end{array}\right]$}} & \multicolumn{1}{c}{\multirow{2}{*}{$\left[\begin{array}{c} D_3=512 \\ N_3=4N \end{array}\right]$}}\\
    & MLP & \multicolumn{5}{c}{} \\ 
    \addlinespace
    \midrule
    \multirow{4}{*}{Stage 4} & DownSample & \multicolumn{5}{c}{[SN]; [Conv 3×3, stride = 1]; [Maxpooling 3×3, stride = 2]} \\
    \cline{2-7}
    \addlinespace
    & SWM & \multicolumn{1}{c}{\multirow{2}{*}{$\left[\begin{array}{c} D_4=256 \\ N_4=N \end{array}\right]$}} & \multicolumn{1}{c}{\multirow{2}{*}{$\left[\begin{array}{c} D_4=384 \\ N_4=N \end{array}\right]$}} & \multicolumn{1}{c}{\multirow{2}{*}{$\left[\begin{array}{c} D_4=384 \\ N_4=N \end{array}\right]$}} & \multicolumn{1}{c}{\multirow{2}{*}{$\left[\begin{array}{c} D_4=512 \\ N_4=N \end{array}\right]$}} & \multicolumn{1}{c}{\multirow{2}{*}{$\left[\begin{array}{c} D_4=512 \\ N_4=N \end{array}\right]$}}\\
    & MLP & \multicolumn{5}{c}{} \\ 
    \addlinespace
    \bottomrule
  \end{tabular}}\\
  \end{center}
  {\scriptsize \hspace{0.128\textwidth} The model sizes of Spikinghash range from Ti to XL. \(D_i\) and \(N_i\) are the embedding dimension and number of patches in stage i,
  
  \hspace{0.128\textwidth} respectively. Conv denotes a convolutional layer with a kernel size \(3 \times 3\) and stride = 1. MP denotes a maxpooling layer with a
  
  \hspace{0.128\textwidth} kernel size \(3 \times 3\) and stride = 2. SN denotes a spiking neuron.}
\end{table*}

\begin{figure}[tb] \centering
  \includegraphics[width=0.35\textwidth,height=0.25\textwidth]{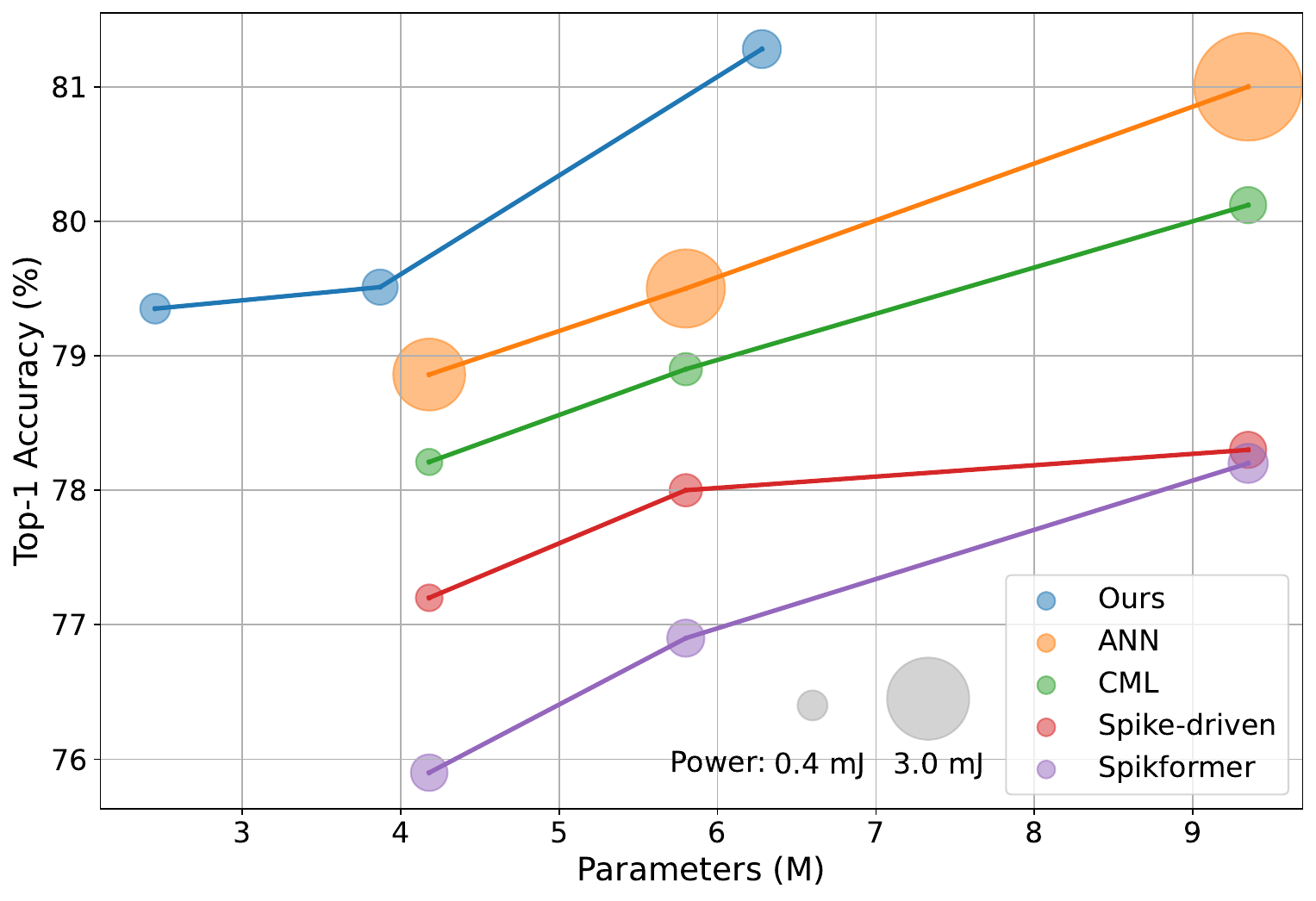}
  \caption{Comparison of the top-1 accuracies on CIFAR100 between Spikinghash and several SNNs \cite{zhou2023enhancing, yao2024spike, zhou2022spikformer} and ANNs \cite{dosovitskiy2020image} (Transformer with four blocks). Power represents the theoretical energy consumption during an evaluation. The bubble size corresponds to energy consumption.} \label{fig:4}
  \vspace{-5pt}
  \label{fig:4}
\end{figure}
\subsection{Theoretical Energy Consumption Analysis}
\label{sec:1}
Energy consumption is a critical metric for evaluating the performance of SNNs. We estimated the energy consumption of Spikinghash according to \cite{yao2023attention, zhou2023spikingformer, yao2024spike}. We calculate the number of synaptic operations of spikes before calculating the theoretical energy consumption for Spikinghash:
\begin{equation}
  {SOP}^l = R \times T \times {FLOPs}^l, \tag{25}
  \label{eq:24}
\end{equation}
where \(l\)  represents a block or layer in Spikinghash, \(R\)  is the firing rate of the block or layer, and \(T\) is the time step of the spike neuron. \({FLOPs}^l\) refers to floating-point operations of block or layer \(l\) . We assume that the multiply and accumulate (MAC) and accumulate (AC) operations are implemented on the 45 nm hardware\cite{horowitz20141}, where $E_{\text{MAC}} = 4.6 \, \text{pj}$ and $E_{\text{AC}} = 0.9 \, \text{pj}$. Spikinghash consists of Spiking Self-Attention (SSA), Spiking WaveMixer (SWM), MLP, and downsample layers. The theoretical energy consumption of a Spikinghash can be calculated as follows:
\begingroup
{\small
\begin{align}
  E_{\text{Spikinghash}} &= E_{\text{AC}} \times \left( \sum_{i=2}^{N} {SOP}_{\text{Conv}}^i + \sum_{j=1}^{M} {SOP}_{\text{SSA}}^j + \sum_{k=1}^{K} {SOP}_{\text{SWM}}^k \right) \nonumber \\
  &\quad + E_{\text{MAC}} \times \left( {FLOP}_{\text{Conv}}^1 \right), \label{eq:26}\tag{26}
\end{align}
}
\endgroup
Eq.~\eqref{eq:26} expresses the overall energy consumption of Spikinghash \({E}_\text{Spikinghash}\). \({FLOP}_\text{Conv}^1\) represents the FLOPs produced by the first Conv layer (MAC operation).

To more comprehensively illustrate the efficiency and performance benefits of Spikinghash, we compare the performance of Spikinghash under different parameters and energy consumption conditions with those of other methods. As shown in Fig.~\ref{fig:4}, Spikinghash achieves a better performance, with energy consumption similar to that of the SNN methods, and significantly lower than that of the ANN method, while maintaining a reduced number of parameters.

\subsection{Hashing Retrieval}
\label{sec:2}
\subsubsection{Experimental Setting}
\paragraph{CIFAR10} CIFAR10 \cite{krizhevsky2009learning} consists of 50,000 training images and 10,000 testing images, with each image having a resolution of \(32 \times 32\) pixels, and includes 10 categories. We conducted experiments via Spikinghash-S and Spikinghash-M (Table~\ref{tab:1}). The timestep is set to 4, and the batch size is set to 64. The parameters \(\alpha\) and \(\beta\) are set to 0.2 and 0.8, respectively. We set the maximum number of epochs to 400 and adopt the Adamp optimizer with a learning rate of 5e-3, which is reduced with cosine decay. We use mAP@5000 for evaluation. As significant inter-class similarity differences exist in CIFAR10, we use the hard similarity loss instead of the dynamic soft similarity loss. 
\paragraph{UCF101-DVS/HMDB51-DVS} UCF101-DVS \cite{bi2020graph} and HMDB51-DVS \cite{bi2020graph} are neuromorphic versions of the UCF101 and HMDB51 datasets, respectively, recorded with a DVS camera under controlled lighting. UCF101-DVS contains 13,320 event streams for 101 types of human actions, while HMDB51-DVS contains 6,766 event streams for 51 types of human actions. For both datasets, we maintain the same experimental settings as \cite{zhou2023spikingformer}, setting the maximum number of epochs to 200 and the time steps to 8. For UCF101-DVS, we employed the official split files of the original UCF101 videos to divide the training and validation sets, while for HMDB51-DVS, we employed a random 7:3 split for the training and test sets. Both datasets were evaluated using \text{mAP@30}. The parameters $\alpha$ and $\beta$ are set to \(1e-3\) and \(1.0\), respectively. The optimizer is AdamP, and the batch size is set to 6. The learning rate is initialized to \(5e-3\) and reduced with cosine decay. We conducted experiments via Spikinghash-M, Spikinghash-L and Spikinghash-XL.
\paragraph{HARDVS} HARDVS \cite{wang2024hardvs} consists of over 100,000 DVS video clips recorded by the DAVIS346 camera. As the first large-scale DVS video dataset of real-world scenes, it covers 300 types of human activities from daily life. The training and test set splits are based on the official partition file, and mAP@100 is used as the evaluation metric. The maximum number of epochs is set to 200, and the time steps is set to 8. The parameters $\alpha$ and $\beta$ are set to \(1e-3\) and \(1.0\), respectively. The optimizer is AdamP, and the batch size is set to 6. The learning rate is initialized to \(5e-3\) and reduced with cosine decay. The experiment is conducted using Spikinghash-XL, as outlined in Table~\ref{tab:1}, with an added maxpooling layer for spatial downsampling. 
\paragraph{Other experimental Setting}For static images, a single image is duplicated \( T \) times as input for multiple time steps. For DVS datasets, the event stream is integrated into multiple frames on the basis of a specified time window.

To update and optimize \(\boldsymbol{S_{\text{soft}}}\), we divide the total training epochs into different rounds. At the start of training, the expressive capacity of features is limited, so we set the first 50 epochs as the initial round and use the binary similarity matrix \(\boldsymbol{M}_{\text{hard}}\) for supervision. With the training progresses, the expressive capacity of features is enhanced, and every 20 epochs are considered one round. In the final 20 epochs of training, each epoch is considered an independent round.

Considering that there is no existing retrieval method in SNNs, we conducted experiments with Spikingformer-CML \cite{zhou2023enhancing} using settings and a hash layer consistent with Spikinghash. Furthermore, there are no existing retrieval methods on UCF101-DVS, HMDB51-DVS and HARDVS and open-source self-supervised video hashing code, we implemented two classic video Transformer models (ANNs) through the publicly available codes, Timesformer \cite{bertasius2021space} and ViViT \cite{arnab2021vivit}, and applied the hash layer of ANNs\cite{wang2023contrastive} to conduct retrieval experiments for comparison with our method.

\subsubsection{Results and Analysis}
\begin{table}[ht]
  \begin{center}
  \caption{mAP comparison of different methods on CIFAR10 dataset with different bits.}
  \label{tab:2}
  \resizebox{0.48\textwidth}{!}{
  \begin{tabular}{lccccccc}
      \toprule
      \multirow{2}{*}{{Method}} & \multirow{2}{*}{\shortstack{{Param} \\ \vspace{0.15em} \\ {(M)}}} & \multirow{2}{*}{\shortstack{{Power} \\ \vspace{0.15em} \\ {(mJ)}}} & \multicolumn{4}{c}{{CIFAR10}} \\
      \cmidrule(lr){4-7}
      & & & {16-bit} & {32-bit} & {48-bit} & {64-bit} \\
      \midrule
      HashNet \cite{cao2017hashnet} & 58.31 & 3.30 & 51.05 & 62.78 & 66.31 & 68.26 \\
      TransHash \cite{chen2022transhash} & - & - & 90.75 & 91.08 & 91.41 & 91.66 \\
      HashFormer \cite{li2022hashformer} & 88.23 & 20.32 & 91.21 & 91.67 & 92.11 & 92.36 \\
      HAAW \cite{pei2023deep} & 18.40 & 17.54 & - & 91.50 & 91.90 & - \\
      Spikingformer-CML* \cite{zhou2023enhancing} & 9.32 & 0.59 & 91.41 & 92.68 & 92.83 & 92.96 \\
      \midrule
      Spikinghash-S & \textbf{3.87} & \textbf{0.56} & 92.01 & 92.86 & 92.96 & 93.01 \\
      Spikinghash-M & 6.24 & 0.66 & \textbf{92.95} & \textbf{94.09} & \textbf{94.10} & \textbf{94.15} \\
      \bottomrule
  \end{tabular}}\\
  \end{center}
  {\scriptsize \hspace{0.01\textwidth} * The results are reproduced through the publicly available code.}
\end{table}
\paragraph{Retrieval Performance on CIFAR10} Table~\ref{tab:2} illustrates the retrieval performance comparison of Spikinghash with existing deep hashing methods on CIFAR10. Spikinghash-M outperforms the other methods at all hash code lengths, achieving the highest average mAP of 93.8\%, which exceeds the performance of Spikingformer-CML by 1.35\%. While maintaining high accuracy, the number of parameters is significantly lower than that of the other models, with only 6.24 million parameters. Spikinghash-S achieves 92.71\% average mAP with only 3.87 M parameters, reducing the number of parameters by 58.2\% compared with Spikingformer-CML. Additionally, Spikinghash significantly reduces energy consumption compared with ANNs. The retrieval results demonstrate the effectiveness of Spikinghash.

\paragraph{Retrieval Performance on UCF101-DVS} Table~\ref{tab:3} illustrates the retrieval performance comparison results of Spikinghash with other models. Spikinghash-XL outperforms the other methods at all hash code lengths, achieving the highest average mAP of 71.66\%, which exceeds the performance of Spikingformer-CML \cite{zhou2023enhancing} by 5.2\%. Spikinghash-XL not only demonstrates outstanding retrieval performance but also has fewer parameters than the other models do, with only 11.3 million parameters. Spikinghash-L also achieves competitive results while maintaining a smaller-scale model size, reducing the number of parameters by 45.8\% compared with Spikingformer-CML. Additionally, in terms of the number of parameters and energy consumption, Spikinghash-M is better than other methods while achieving higher retrieval performance. These results demonstrate the efficacy of Spikinghash in retrieval task.
\begin{table}[ht]
  \begin{center}
  \caption{mAP comparison of different methods on UCF101-DVS dataset with different bits.}
  \label{tab:3}
  \resizebox{0.48\textwidth}{!}{
  \begin{tabular}{lccccccc}
      \toprule
      \multirow{2}{*}{{Method}} & \multirow{2}{*}{\shortstack{{Param} \\ \vspace{0.15em} \\ {(M)}}} & \multirow{2}{*}{\shortstack{{Power} \\ \vspace{0.15em} \\ {(mJ)}}} & \multicolumn{3}{c}{{UCF101-DVS}} \\
      \cmidrule(lr){4-6}
      & & & {64-bit} & {128-bit} & {256-bit} \\
      \midrule
      ViViT* \cite{arnab2021vivit}	& 26.10	& 60.52	& 47.32	& 47.75	& 48.26 \\
      Timesformer* \cite{bertasius2021space}	& 18.20	& 83.30 & 51.75	& 54.28	& 52.25\\
      Spikingformer-CML* \cite{zhou2023enhancing} & 16.60 & 9.13	& 63.69	& 67.03	& 68.66\\
      \midrule
      Spikinghash-M	& \textbf{6.50}	& \textbf{3.56}	& 67.61	& 68.52	& 69.73\\
      Spikinghash-L	& 9.03 & 10.12	& 68.74	& 71.61	& 73.24\\
      Spikinghash-XL & 11.30	& 12.31	& \textbf{69.28}	& \textbf{72.33}	& \textbf{73.36}\\
      \bottomrule
  \end{tabular}}\\
  \end{center}
  {\scriptsize * The results are reproduced through the publicly available code.}
\end{table}
\paragraph{Retrieval Performance on HMDB51-DVS} Table~\ref{tab:4} shows the mAP comparison results of Spikinghash with other models on HMDB51-DVS. Spikinghash-L outperforms the other methods at all hash code lengths, achieving the highest average mAP of 54.13\%, which exceeds the performance of Spikingformer-CML by 2.21\%. Spikinghash-L not only demonstrates outstanding retrieval performance but also has fewer parameters than other models do. Spikinghash-M also achieves competitive results while maintaining a smaller-scale model size, reducing the number of parameters by 60.8\% compared with Spikingformer-CML.

\begin{table}[ht]
  \begin{center}
  \caption{mAP comparison of different methods on HMDB51-DVS dataset with different bits.}
  \label{tab:4}
  \resizebox{0.48\textwidth}{!}{
  \begin{tabular}{lcccccc}
      \toprule
      \multirow{2}{*}{{Method}} & \multirow{2}{*}{\shortstack{{Param} \\ \vspace{0.15em} \\ {(M)}}} & \multirow{2}{*}{\shortstack{{Power} \\ \vspace{0.15em} \\ {(mJ)}}} & \multicolumn{3}{c}{{HMDB51-DVS}} \\
      \cmidrule(lr){4-6}
      & & & {64-bit} & {128-bit} & {256-bit} \\
      \midrule
      ViViT* \cite{arnab2021vivit} & 26.10 & 60.50 & 25.75 & 25.83 & 26.54 \\
      Timesformer* \cite{bertasius2021space} & 18.20 & 83.3 & 41.92 & 43.26 & 41.48 \\
      Spikingformer-CML* \cite{zhou2023enhancing} & 16.60 & 7.90 & 51.63 & 51.79 & 52.32 \\
      \midrule
      Spikinghash-M & \textbf{6.50} & \textbf{2.47} & 51.72 & 51.92 & 52.61 \\
      Spikinghash-L & 9.03 & 8.93 & \textbf{52.47} & \textbf{54.00} & \textbf{55.91} \\
      \bottomrule
  \end{tabular}}\\
  \end{center}
  {\scriptsize * The results are reproduced through the publicly available code.}
\end{table}

\paragraph{Retrieval Performance on HARDVS} As a large-scale DVS dataset, HARDVS offers a more rigorous and comprehensive benchmark for evaluating the effectiveness of models. Table~\ref{tab:5} shows the mAP comparison results of Spikinghash with other models. Spikinghash-XL outperforms the other methods at all hash code lengths, achieving the highest average mAP of 48.39\%. Spikinghash not only demonstrates outstanding retrieval performance but also has fewer parameters than other models do.

\begin{table}[ht]
  \begin{center}
  \caption{mAP comparison of different methods on HARDVS dataset with different bits.}
  \label{tab:5}
  \resizebox{0.43\textwidth}{!}{
  \begin{tabular}{lccccc}
      \toprule
      \multirow{2}{*}{{Method}} & \multirow{2}{*}{\shortstack{{Param} \\ \vspace{0.15em} \\ {(M)}}} & \multicolumn{3}{c}{{HARDVS}} \\
      \cmidrule(lr){3-5}
      & & {64-bit} & {128-bit} & {256-bit} \\
      \midrule
      Timesformer* \cite{bertasius2021space} & 18.20 & 41.80 & 42.20 & 44.56 \\
      Spikingformer-CML* \cite{zhou2023enhancing} & 16.60 & 43.64 & 49.13 & 49.33 \\
      \midrule
      Spikinghash-XL & \textbf{11.30} & \textbf{45.39} & \textbf{49.51} & \textbf{50.27} \\
      \bottomrule
  \end{tabular}}\\
  \end{center}
  {\scriptsize \hspace{0.03\textwidth} * The results are reproduced through the publicly available code.}
\end{table}
\vspace{-3.5mm}
\subsection{Event-based Action Recognition}
\label{sec:3}
While demonstrating outstanding retrieval performance, Spikinghash optimizes temporal dependencies and local spatial feature extraction capabilities in SNN-Transformers, enabling it to perform well in other tasks. In the following sections, we conduct experiments on action recognition and image classification to validate the superior feature extraction capabilities of Spikinghash. 
\subsubsection{Experimental Setting}
\paragraph{CIFAR10-DVS} CIFAR10-DVS \cite{li2017cifar10} is the neuromorphic version of CIFAR10, where 10,000 original images are converted into event streams via the DVS camera. These event streams are divided into 16 windows, and the data are split into training and test sets at a 9:1 ratio following the preprocessing method in \cite{zhou2023spikingformer}. We use the AdamP optimizer with a learning rate of \(5e-3\) over 100 epochs, which is reduced via cosine decay. The batch size is set to 16. We conducted experiments using Spikinghash-Ti, as outlined in Table~\ref{tab:1}.
\paragraph{DVS128 Gesture} DVS128 Gesture \cite{amir2017low} is captured directly via a DVS camera and contains 1,342 gesture action sequences, categorized into 11 classes, with an input resolution of \(128 \times 128\). We use the same preprocessing method in \cite{zhou2023spikingformer} and split the data into training and validation sets at an 8:2 ratio. The time step of the spiking neuron is 16, and the batch size is set to 16. We use the AdamW optimizer with a learning rate of \(5e-4\) over 200 epochs, which is reduced via cosine decay. The batch size is set to 16. We conducted experiments using Spikinghash-Ti, as outlined in Table~\ref{tab:1}.
\paragraph{Other experimental Setting} For UCF101-DVS, HMDB51-DVS and HARDVS, the experimental settings remain consistent with those in the retrieval experiment. In action recognition and image classification experiments, we optimized the model using only the classification loss \(L_{\mathrm{cls}}\).
\subsubsection{Results and Analysis}
\paragraph{Performance on HARDVS} The metrics of Spikinghash on HARDVS are compared with the current SOTA methods in Table~\ref{tab:6}. Spikinghash-XL achieves the top-1 accuracy of 51.1\%, surpassing the previous SOTA SNN method by 3.63\%. Spikinghash not only demonstrates outstanding performance but also has fewer parameters than other models, with only 11.3 million parameters. Spikinghash demonstrates competitive accuracy compared to ANN models while achieving significantly lower energy consumption. Spikinghash-L also achieves better results than the other SNN model while maintaining a smaller-scale model size, achieving the top-1 accuracy of 50.1\%. By directly using the classification head after the hash layer, Spikinghash-XL (Hash) achieves the top-1 accuracy of 51.0\%. These results not only highlight the exceptional temporal modeling capabilities of Spikinghash but also emphasize its superior efficacy on DVS datasets.
\begin{table}[ht]
  \begin{center}
  \caption{Comparison with previous works on HARDVS.}
  \label{tab:6}
  \vspace{-1mm}
  \resizebox{0.49\textwidth}{!}{
  \begin{tabular}{lcccccc}
      \toprule
      \multirow{2}{*}{{Methods}} & \multirow{2}{*}{{Type}} & \multirow{2}{*}{{Architecture}} & \multirow{2}{*}{\shortstack{{Power} \\ \vspace{0.15em} \\ {(mJ)}}} & \multirow{2}{*}{\shortstack{{Param} \\ \vspace{0.15em} \\ {(M)}}} & \multirow{2}{*}{\shortstack{{Time} \\ \vspace{0.15em} \\ {Step}}} & \multirow{2}{*}{\shortstack{{Acc} \\ \vspace{0.15em} \\ {(\%)}}}\\ \\
      \midrule
      SlowFast \cite{feichtenhofer2019slowfast} & ANN & - & - & 33.6 & 8 & 46.5 \\
      ACTION-Net \cite{wang2021action} & ANN & - & - & 27.9 & 8 & 46.9 \\
      TSM \cite{lin2019tsm} & ANN & ResNet50 & 87.4 & 24.3 & 8 & \textbf{52.6} \\
      Timesformer \cite{bertasius2021space}& ANN & VIT & - & 121.2 & 8 & 50.8 \\
      ESTF \cite{wang2024hardvs} & ANN & Resnet18 & 81.1 & 46.7 & 8 & 51.2 \\
      \midrule
      Spike-Driven V2\cite{yao2024spike2} & SNN & Meta-SpikeFormer & - & 18.3 & 8 & 47.5 \\
      \midrule
      \multirow{3}{*}{Spikinghash (Ours)} & SNN & Spikinghash-L  & \textbf{12.8} & \textbf{9.0}  & 8 & 50.1  \\
       & SNN & Spikinghash-XL (Hash)  & 13.4 & 11.3  & 8 & 51.0 \\
       & SNN & Spikinghash-XL & 13.4 & 11.3 & 8 & \textbf{51.1} \\
      \bottomrule
  \end{tabular}}\\
  \end{center}
  {\scriptsize Spikinghash-XL (Hash) is a classification method which from the output of the hash layer.}
\end{table}
\paragraph{Performance on CIFAR10-DVS/DVS128 Gesture} Table~\ref{tab:7} shows the accuracy comparison of Spikinghash against other methods on CIFAR10-DVS and DVS128 Gesture. For CIFAR10-DVS, Spikinghash outperforms most of the SNNs methods, achieving the top-1 accuracy of 82.9\%, which is 1.5\% higher than that of Spikingformer-CML. For DVS128 Gesture, Spikinghash achieves the same accuracy as the current SOTA model \cite{yao2024spike}, attaining the top-1 accuracy of 99.3\%.
\vspace{-2mm}
\begin{table}[ht]
  \centering
  \caption{Comparison with previous works on CIFAR10-DVS and DVS128 Gesture.}
  \label{tab:7}
  \resizebox{0.48\textwidth}{!}{
  \begin{tabular}{lccccc}
      \toprule
      \multirow{2}{*}{{Method}} & \multicolumn{2}{c}{{CIFAR10-DVS}} & \multicolumn{2}{c}{{DVS128-Gesture}} \\
      \cmidrule(lr){2-3} \cmidrule(lr){4-5}
      & {Time Step} & {Acc(\%)} & {Time Step} & {Acc(\%)} \\
      \midrule
      STBP-tdBN \cite{zheng2021going} & 10 & 67.8 & 40 & 96.9 \\
      PLIF \cite{fang2021incorporating} & 20 & 74.8 & 20 & 97.6 \\
      Spikformer \cite{zhou2022spikformer} & 16 & 80.6 & 16 & 97.9 \\
      Spikingformer \cite{zhou2023spikingformer} & 16 & 81.3 & 16 & 98.3 \\
      Spikingformer-CML \cite{zhou2023enhancing} & 16 & 81.4 & 16 & 98.6 \\
      Spike-driven Transformer \cite{yao2024spike} & 16 & 80.0 & 16 & 99.3 \\
      STS-Transformer \cite{wang2023spatial} & 16 & 79.9 & 16 & 98.7 \\
      SpikingResformer \cite{shi2024spikingresformer} & - & - & 16 & 98.6 \\
      QKFormer \cite{zhou2024qkformer} & 16 & \textbf{84.0} & 16 & 98.6 \\
      \midrule
      Spikinghash-Ti & 16 & 82.9 & 16 & \textbf{99.3} \\
      \bottomrule
  \end{tabular}}
\end{table}
\paragraph{Performance on UCF101-DVS/HMDB51-DVS} The metrics of Spikinghash on UCF101-DVS and HMDB51-DVS are compared with the current methods in Table~\ref{tab:8}. Spikinghash achieves the best performance on both datasets. Specifically, Spikinghash-XL achieves the top-1 accuracy of 72.1\% on UCF101-DVS, surpassing the SOTA method SpikePoint \cite{ren2023spikepoint} by 3.6\%. For HMDB51-DVS, Spikinghash-XL achieves the top-1 accuracy of 59.1\%, which exceeds the performance of SpikePoint by 3.5\%. Spikinghash-L also achieves better results than the other models while maintaining a smaller-scale model size, achieving the top-1 accuracy of 57.6\% on HMDB51-DVS and 71.7\% on UCF101-DVS.
\begin{table}[ht]
  \begin{center}
  \caption{Comparison with previous works on UCF101-DVS and HMDB51-DVS.}
  \label{tab:8}
  \resizebox{0.48\textwidth}{!}{
  \begin{tabular}{lcccccc}
      \toprule
      \multirow{2}{*}{{Method}} & \multirow{2}{*}{{Type}} & \multicolumn{2}{c}{{HMDB51-DVS}} & \multicolumn{2}{c}{{UCF101-DVS}} \\
      \cmidrule(lr){3-4} \cmidrule(lr){5-6}
      & & {Time Step} & {Acc(\%)} & {Time Step} & {Acc(\%)} \\
      \midrule
      RG-CNN+Res.3D \cite{bi2020graph} & ANN & $t = 240$ & - & $t = 240$ & 63.2 \\
      RG-CNN+Incep.3D \cite{bi2020graph} & ANN & $t = 240$ & 45.5 & - & - \\
      I3D \cite{bi2020graph} & ANN & $t = 240$ & 38.6 & - & - \\
      Res-SNN-18 \cite{fang2021deep} & SNN & 8 & 42.6 & 8 & 57.8 \\
      RM-Res-SNN-18 \cite{yao2023sparser} & SNN & 8 & 44.7 & 8 & 58.5 \\
      SpikePoint \cite{ren2023spikepoint} & SNN & - & 55.6 & - & 68.5 \\
      \midrule
      Spikinghash-L & SNN & 8 & 57.6 & 8 & 71.7 \\
      Spikinghash-XL & SNN & 8 & \textbf{59.1} & 8 & \textbf{72.1} \\
      \bottomrule
  \end{tabular}}\\
  \end{center}
  {\scriptsize \hspace{0.01\textwidth} \(t\) represents the time duration of each sample input, in milliseconds.}
\end{table}
\vspace{-5mm}
\subsection{Static Image Classification}
\label{sec:4}
\subsubsection{Experimental Setting}
\paragraph{CIFAR10/CIFAR100} The difference between CIFAR10 and CIFAR100 lies in the number of categories they contain: CIFAR10 includes 10 categories, whereas CIFAR100 has 100 categories, making the classification more challenging. The experimental settings are consistent with those used in the CIFAR10 retrieval experiment. We conducted experiments via Spikinghash-S and Spikinghash-M (Table~\ref{tab:1}).

\paragraph{Tiny-ImageNet} Tiny-ImageNet is a subset of ImageNet that includes images from 200 categories. Each category consists of 500 training images, 50 validation images, and 50 test images, with each image having a resolution of \(64 \times 64\) pixels. We conducted experiments via Spikinghash-M, as outlined in Table~\ref{tab:1}, with an added maxpooling layer for spatial downsampling. The time step is set to 4, and the batch size is set to 16. We set the maximum number of epochs to 310 and adopt the Adamw optimizer with a learning rate of \(5e-3\), which is reduced with cosine decay.
\subsubsection{Results and Analysis}
\begin{table}[ht]
  \begin{center}
  \caption{Comparison with previous works on CIFAR10 and CIFAR100.}
  \label{tab:9}
  \resizebox{0.48\textwidth}{!}{
  \begin{tabular}{lcccc}
      \toprule
      \multirow{2}{*}{{Methods}} & \multirow{2}{*}{\shortstack{{Param} \\ \vspace{0.15em} \\ {(M)}}} & \multirow{2}{*}{\shortstack{{Time} \\ \vspace{0.15em} \\ {Step}}} & \multicolumn{2}{c}{{Acc (\%)}} \\
      \cmidrule(lr){4-5}
      & & & {CIFAR10} & {CIFAR100} \\
      \midrule
      STBP-tdBN \cite{zheng2021going} & 12.63 & 6 & 93.16 & - \\
      PLIF \cite{fang2021incorporating} & - & 8 & 93.50 & - \\
      Dspike \cite{li2021differentiable} & 11.17 & 6 & 94.25 & 74.24 \\
      Spikformer \cite{zhou2022spikformer} & 9.32 & 4 & 95.19 & 77.86 \\
      Spikingformer \cite{zhou2023spikingformer} & 9.32 & 4 & 95.61 & 79.09 \\
      Spike-driven Transformer \cite{yao2024spike} & 9.32 & 4 & 95.60 & 78.40 \\
      Spikingformer-CML \cite{zhou2023enhancing} & 9.32 & 4 & 95.95 & 80.37 \\
      SpikingResformer-Ti \cite{shi2024spikingresformer} & 10.79 & 4 & 96.24 & 79.28 \\
      QKFormer \cite{zhou2024qkformer} & 6.74 & 4 & 96.18 & 81.15 \\
      \midrule
      Spikinghash-S & \textbf{3.87} & 4 & 96.01 & 79.51 \\
      Spikinghash-M & 6.24 & 4 & \textbf{96.50} & \textbf{81.28} \\
      \bottomrule
  \end{tabular}}\\
  \end{center}
\end{table}
\paragraph{Performance on CIFAR10/CIFAR100} Table~\ref{tab:9} illustrates the accuracy comparison of Spikinghash with other models on CIFAR10 and CIFAR100. For CIFAR10, Spikinghash-M significantly outperforms all the other models, achieving the top-1 accuracy of 96.50\%. Spikinghash-M not only maintains high accuracy but also further reduces the number of parameters by 33.5\% compared with Spikingformer-CML \cite{zhou2023enhancing}. Spikinghash-S also achieves better results than the other models while maintaining a smaller-scale model size, achieving the top-1 accuracy of 96.01\%. For CIFAR100, Spikinghash-M demonstrates superior performance, achieving the top-1 accuracy of 81.28\%. Spikinghash-S achieves the top-1 accuracy of 79.51\% with only 3.87 million parameters, reducing the parameters by 58.5\% compared with Spikingformer-CML.

\paragraph{Performance on Tiny-ImageNet} Table~\ref{tab:10} illustrates the accuracy comparison of Spikinghash with other methods on Tiny-ImageNet. Spikinghash-M outperforms all the other methods, achieving the top-1 accuracy of 68.26\%. Additionally, while maintaining high accuracy, Spikinghash-M further reduces the number of parameters, resulting in a 25.3\% decrease compared with Spikingformer-CML \cite{zhou2023enhancing}. These results convincingly demonstrate the effectiveness of Spikinghash.
\begin{table}[ht]
  \begin{center}
  \caption{Performance of Spikinghash compared with existing methods on Tiny-ImageNet.}
  \label{tab:10}
  \resizebox{0.45\textwidth}{!}{
  \begin{tabular}{lcccc}
      \toprule
      \multirow{2}{*}{{Methods}} & \multirow{2}{*}{{Architecture}} & \multirow{2}{*}{\shortstack{{Param} \\ \vspace{0.15em} \\ {(M)}}} & \multirow{2}{*}{\shortstack{{Time} \\ \vspace{0.15em} \\ {Step}}} & \multirow{2}{*}{\shortstack{{Acc} \\ \vspace{0.15em} \\ {(\%)}}}\\ \\
      \midrule
      % DCT \cite{garg2021dct} & VGG-13 & 60.56 & 125 & 56.90 \\
      QCFS \cite{bu2023optimal} & VGG-16 & 65.87 & 32 & 53.54 \\
      Online-LTL \cite{yang2022training} & VGG-13 & 60.56 & 16 & 54.82 \\
      Offline-LT \cite{yang2022training} & VGG-13 & 60.56 & 16 & 55.37 \\
      ASGL \cite{wang2023adaptive} & VGG-13 & 60.56 & 8 & 56.81 \\
      Spikformer* \cite{zhou2022spikformer} & Spikformer & 9.40 & 4 & 65.70 \\
      Spikingformer-CML* \cite{zhou2023enhancing} & Spikingformer & 8.40 & 4 & 66.59 \\
      \midrule
      Ours & Spikinghash-M & \textbf{6.28} & 4 & \textbf{68.26} \\
      \bottomrule
  \end{tabular}}\\
  \end{center}
  {\scriptsize \hspace{0.02\textwidth} * The results are reproduced through the publicly available code.}
\end{table}
\vspace{-2mm}
\subsection{Ablation Study}
\label{sec:5}
\subsubsection{Temporal Modeling Efficacy}
\label{sec:6}
We select action pairs from HMDB51-DVS that contain symmetric semantic information at different time steps, such as ``sit/stand" and ``catch/shoot ball", for analysis. Distinguishing these actions requires recognizing the temporal order and dependencies across multiple time steps. Additionally, we select action pairs that contain temporal dynamics are asymmetrical, such as “hit/climb” and “drink/hug”, for comparative analysis. These action pairs can be distinguished solely on the basis of spatial information.

We use the Spikingformer-CML \cite{zhou2023enhancing} as the baseline for comparison, as it lacks consideration for temporal dependencies. Conversely, our proposed Spikinghash considers the dependencies between different time steps in the SWM, capturing and utilizing the temporal information more effectively. To further verify the performance improvements brought by the SWM due to its specialized design for the time steps in SNNs, we replace the 3D-DWT in SWM with the 2D-DWT, and remove the time step dimension from the learnable parameter matrices, naming the revised model Spikinghash (2D-SWM). In the experiment, we trained each action pair for 50 epochs via Spikinghash-XL, as outlined in Table~\ref{tab:1}, and compare the classification accuracy.

As shown in Table~\ref{tab:11}, the experimental results indicate that for action pairs where the temporal dynamics are symmetrical, Spikinghash achieves greater accuracy than Spikingformer-CML \cite{zhou2023enhancing} and Spikinghash (2D-SWM). For the actions ``sit/stand", Spikinghash outperforms Spikingformer-CML and Spikinghash (2D-SWM) by 11.29\% and 5.67\%, respectively. For the actions ``catch/shoot ball”, Spikinghash outperforms Spikingformer-CML and Spikinghash (2D-SWM) by 9.30\% and 6.98\%, respectively. Conversely, for action pairs where the temporal dynamics are asymmetrical, the accuracies of Spikingformer-CML, Spikinghash, and Spikinghash (2D-SWM) are close. Comparing the accuracy differences between these two types of data effectively demonstrates the ability of Spikinghash to model temporal dependencies.
%\vspace{-4.5mm}
\begin{table}[ht]
  \centering
  \caption{Ablation study results on temporal modeling efficacy.}
  \label{tab:11}
  \resizebox{0.44\textwidth}{!}{
  \begin{tabular}{lccc}
      \toprule
      \multirow{2}{*}{Datasets} & \multicolumn{1}{c}{Spikingformer} & \multicolumn{1}{c}{Spikinghash} & \multicolumn{1}{c}{Spikinghash}\\
      & \multicolumn{1}{c}{-CML} & \multicolumn{1}{c}{(2D-SWM)} & \\
      \midrule
      \text{Sit \& Stand} & 71.43 & 76.47 & \textbf{82.14}\\
      \text{Catch \& Shoot ball} & 86.05 & 88.37 & \textbf{95.35}\\
      \text{Drink \& Hug} & 92.73 & \textbf{94.55} & \textbf{94.55}\\
      \text{Hit \& Climb} & 95.65 & 95.65 & \textbf{96.65}\\
      \bottomrule
  \end{tabular}}
\end{table}
%\vspace{-3mm}
\subsubsection{Ablation study on SWM and SSA}
\label{sec:7}
To compare the effectiveness of the Spiking WaveMixer (SWM) with that of the Spiking Self-Attention (SSA) in shallow layers, we replace the SWM in Spikinghash with SSA, which is named Spikinghash $(\text{all-SSA})$. We conduct classification experiments via the Spikinghash-M, as outlined in Table~\ref{tab:1}. As shown in Table~\ref{tab:12}, for CIFAR10, Spikinghash $(\text{all-SSA})$ exhibits a decrease of 2.30\% compared with Spikinghash, and a decrease of 9.72\% on Tiny ImageNet. These results demonstrate the efficacy of SWM in shallow layers.
\vspace{-3mm}
\begin{table}[ht]
  \centering
  \caption{Ablation study results on SWM and SSA.}
  \label{tab:12}
  \resizebox{0.32\textwidth}{!}{
  \begin{tabular}{lcc}
      \toprule
      \multirow{2}{*}{{Method}} & \multicolumn{2}{c}{{Acc(\%)}} \\
      \cmidrule(lr){2-3}
      & {CIFAR10} & {Tiny-ImageNet} \\
      \midrule
      Spikinghash $(\text{all-SSA})$ & 94.20 & 58.54 \\
      Spikinghash & \textbf{96.50} & \textbf{68.26} \\
      \bottomrule
  \end{tabular}}
\end{table}
%\vspace{-3mm}
\subsubsection{Ablation study on dynamic soft similarity loss}
\label{sec:8}
To verify the effectiveness of the proposed dynamic soft similarity loss \( L_s \), we replace the loss function \( L_s \) in Spikinghash with the conventional similarity loss \( L_h \) (replacing the soft similarity matrix \(\boldsymbol{M}_{\text{soft}}\) with the hard similarity matrix \(\boldsymbol{M}_{\text{hard}}\)) and compare the retrieval performance of both methods on UCF101-DVS. Additionally, we introduce two metrics, Normalized Discounted Cumulative Gain (NDCG) and Average Cumulative Gain (ACG), to further validate the effectiveness of the dynamic soft similarity loss in constraining the feature distances between similar classes. The formulas are as follows:
\begin{align}
  \mathit{ACG}@n &= \frac{1}{n} \sum_{i=1}^{n} r(x_{\text{q}}, x_i), \tag{27} \\
  \mathit{DCG}@n &= r(x_{\text{q}}, x_1) + \sum_{i=2}^{n} \frac{r(x_{\text{q}}, x_i)}{\log_2(i)}, \tag{28} \\
  \mathit{NDCG}@n &= \frac{\mathit{DCG}@n}{\mathit{IDCG}@n}, \tag{29}
\end{align}
where \(\mathit{ACG}@n\) represents the average similarity between the query sample \(x_q\) and the top-\(n\) returned samples. Compared to ACG, NDCG applies a weighting based on the ranking order. \( r(\cdot) \) represents the relevance score of the samples we defined. For samples of the same class, \( r(x_i, x_j) = 1 \); for samples of similar classes, \( r(x_i, x_j) = 0.5 \); for samples of different classes, \( r(x_i, x_j) = 0 \). \( \mathit{IDCG}@n \) denotes the \( \mathit{DCG}@n \) score in the ideal ordering of the returned samples, which constrains the value of \( \mathit{NDCG}@n \) to the range \([0,1]\). As UCF101-DVS does not have an official definition of similar classes, we use the Pre-trained VideoMAE \cite{tong2022videomae} to extract features from UCF101 and perform clustering to identify similar classes, which are then used as ground truth.
\begin{figure*}[tb]
  \centering
  \begin{minipage}{0.45\textwidth}
    \centering
    \includegraphics[width=\textwidth]{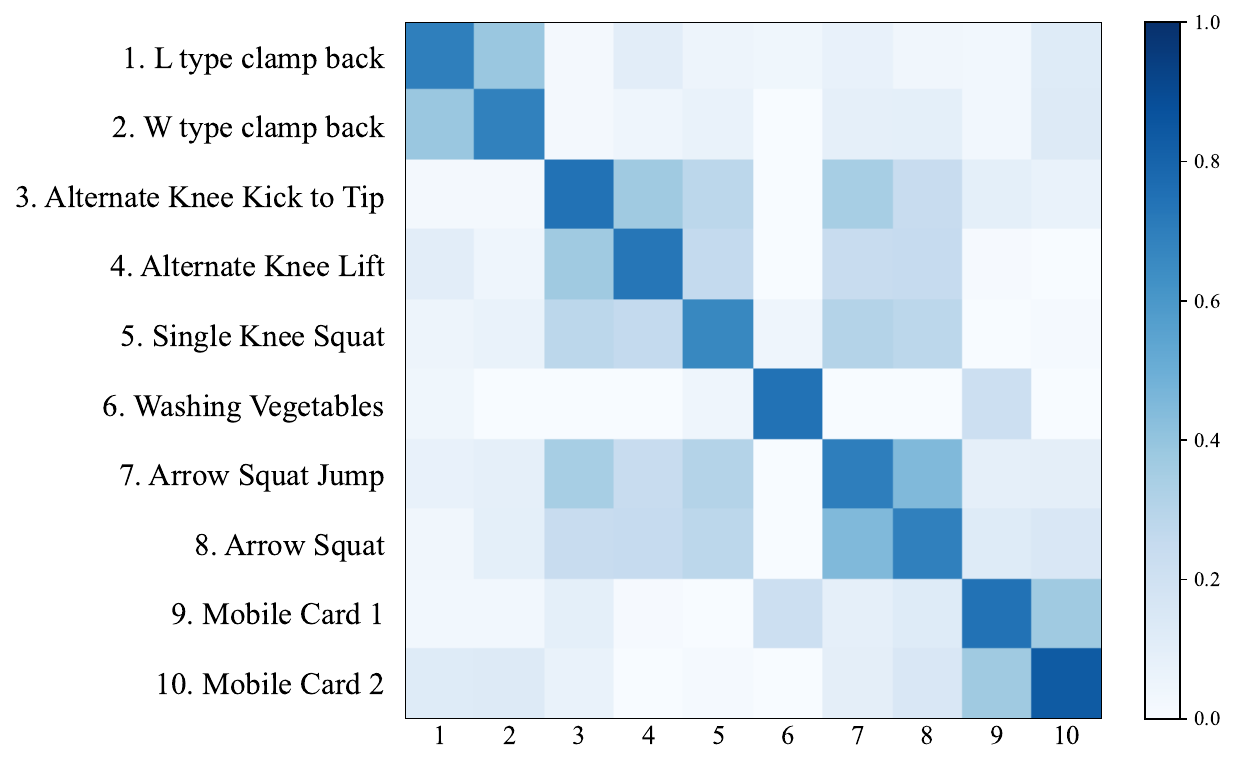}
    {\small \text{\hspace{8em}\(\boldsymbol{S_{\text{soft}}}\)}}
    % \caption{(a)}
    \label{fig:5a}
  \end{minipage}
  \hspace{14pt}
  \begin{minipage}{0.45\textwidth}
    \centering
    \includegraphics[width=\textwidth]{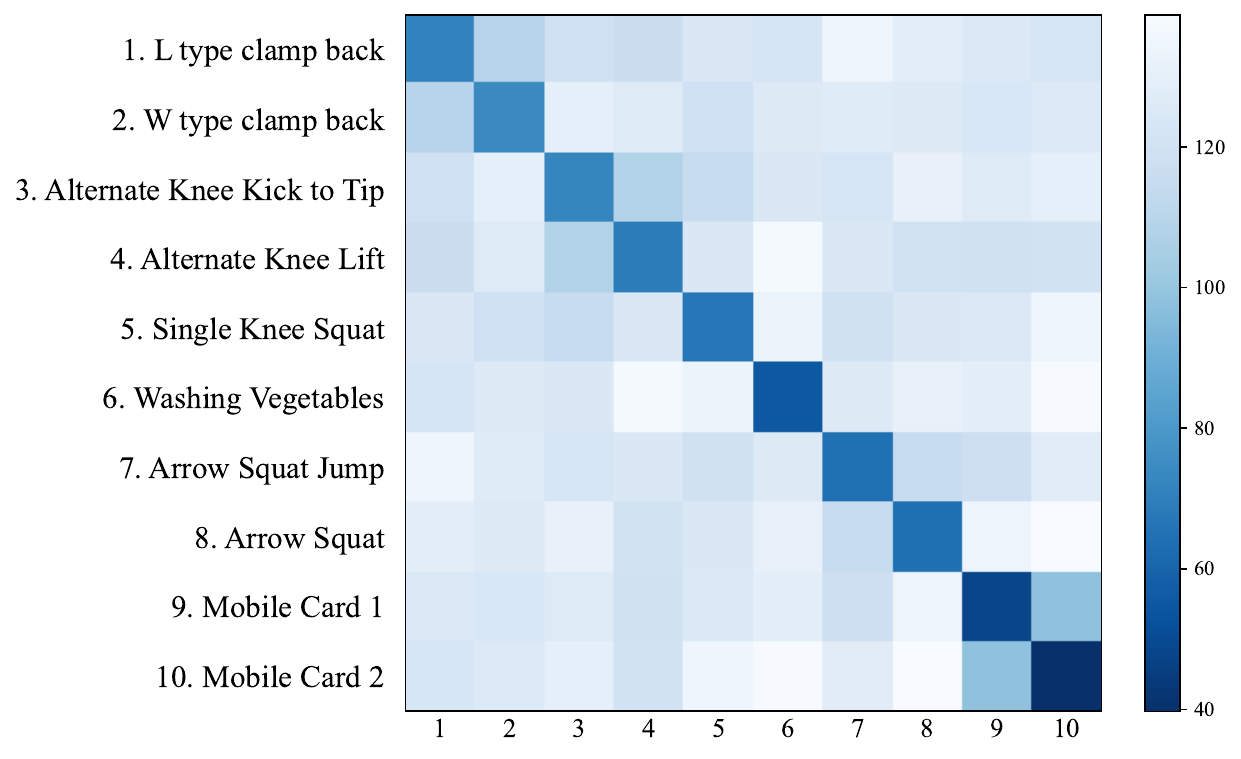}
    {\small \text{\hspace{8em}\(\boldsymbol{S_{\text{hash}}}\)}}
    \label{fig:5b}
  \end{minipage}
  \vspace{-5pt}
  \caption{Visualization of similarity matrixs \(\boldsymbol{S_{\text{soft}}}\) based on membrane potentials and \(\boldsymbol{S_{\text{hash}}}\) based on Hamming distances.}
  \label{fig:5}
\end{figure*}
\begin{table*}[ht]
  \centering
  \caption{Ablation study results of dynamic soft similarity loss on UCF101-DVS.}
  \label{tab:13}
  \resizebox{0.70\textwidth}{!}{
  \begin{tabular}{lccccccccc}
      \toprule
      \multirow{2}{*}{Loss} & \multicolumn{3}{c}{mAP@100} & \multicolumn{3}{c}{ACG@100} & \multicolumn{3}{c}{NDCG@100}\\
      \cmidrule(lr){2-4} \cmidrule(lr){5-7} \cmidrule(lr){8-10}
      & 64-bits & 128-bits & 256-bits & 64-bits & 128-bits & 256-bits & 64-bits & 128-bits & 256-bits\\
      \midrule
      \multirow{1}{*}{\(L_h\) \& \(L_{cls}\)}& 0.672 & 0.679 & 0.719 & 0.597 & 0.610 & 0.646 & 0.653 & 0.666 & 0.703 \\
      \midrule
      \multirow{1}{*}{\(L_s\) \& \(L_{cls}\)}& \textbf{0.682} & \textbf{0.714} & \textbf{0.734} & \textbf{0.615} & \textbf{0.650} & \textbf{0.669} & \textbf{0.671} & \textbf{0.706} & \textbf{0.727}\\
      \bottomrule
  \end{tabular}}
\end{table*}
As shown in Table~\ref{tab:13}, the proposed loss effectively improves the retrieval performance of Spikinghash. Our proposed loss \( L_s \) achieves an improvement of 2.0\% in the average mAP compared to the loss \( L_h \), while the average \( \text{DCG}@100 \) and \( \text{NDCG}@100 \) are enhanced by 0.0270 and 0.0273, respectively. These results effectively demonstrate that the dynamic soft similarity loss can capture more detailed inter-class similarity differences and enhance retrieval performance.

We visualize the soft similarity matrix \(\boldsymbol{S}_{\text{soft}}\) learned from the retrieval experiments on HARDVS, and the similarity matrix \(\boldsymbol{S}_{\text{hash}}\) computed from the Hamming distances between hash codes. As shown in Fig~\ref{fig:5}, we selected ten categories for visualization, with both the soft similarity and Hamming distances represented by varying color intensities. The two matrices exhibit similar visual patterns. The darkest shades along the diagonal indicate that the Spikinghash learned the highest similarity (the smallest Hamming distance) for samples within the same category. Moreover, actions with similar characteristics, such as ``L-type clamp back” and ``W-type clamp back,” as well as ``Mobile Card 1” and ``Mobile Card 2,” exhibit a high degree of similarity (low Hamming distance). Conversly, for some different categories, such as ``Washing Vegetables,” show lower similarity (larger Hamming distance) with other actions. The difference between two similarity matrices results from the fact that hash codes are binary vectors after dimensionality reduction and quantization, and they are constrained by classification loss. By employing the similarity matrix \(\boldsymbol{S_{\text{soft}}}\) as soft labels, Spikinghash can capture finer-grained similarities between categories, leading to improved retrieval performance. The whole soft similarity matrix \(\boldsymbol{S_{\text{soft}}}\) for all categories are provided in the supplementary materials.

\section{Conclusion}
\label{sec:conclusion}
In this work, we propose a novel supervised hashing method named Spikinghash with a hierarchical lightweight structure. Based on the binary characteristics of SNNs, Spikinghash achieves efficient retrieval for DVS data. By deploying Spiking WaveMixer (SWM) in shallow layers, Spikinghash can effectively overcome the current limitations of SNN-Transformers. SWM uses multilevel 3D-DWT to decouple features into low-frequency and high-frequency components and performs spatiotemporal spectral feature fusion. SWM can effectively extract local spatial features and capture the dependencies among multiple time steps. Spiking Self-Attention (SSA) is deployed in deeper layers to further extract global spatiotemporal features, effectively integrating the fine-grained features learned by SWM. Furthermore, a dynamic soft similarity loss is proposed to leverage inter-class similarity differences and enhancing retrieval performance. Extensive experiments across various datasets show that the Spikinghash achieves state-of-the-art results with low energy consumption and fewer parameters.

{
    \small
    \bibliographystyle{IEEEtran}
    \bibliography{7_references}
}

\end{document}